\documentclass[sigconf]{acmart}
\usepackage{multirow}
\usepackage{tabularx}
\usepackage{graphicx}
\usepackage{longtable}
\usepackage{fancyhdr}

\AtBeginDocument{%
  \providecommand\BibTeX{{%
    \normalfont B\kern-0.5em{\scshape i\kern-0.25em b}\kern-0.8em\TeX}}}

\renewcommand\footnotetextcopyrightpermission[1]{}
\setcopyright{none}

\acmBooktitle{Woodstock '18: ACM Symposium on Neural Gaze Detection,
 June 03--05, 2018, Woodstock, NY} 
\begin{document}

\title{OlaGPT: Empowering LLMs With Human-like Problem-Solving Abilities}


\author{Yuanzhen Xie, Tao Xie, Mingxiong Lin, WenTao Wei, Chenglin Li, Beibei Kong, \\ Lei Chen, Chengxiang Zhuo, Bo Hu, Zang Li}
\affiliation{%
  \institution{Platform and Content Group, Tencent}
  \country{Shenzhen, Guangdong, China}}
\email{{xieyzh3, taoxie168}@gmail.com}
\email{{matrixmxlin, lubewtwei, chainli, echokong, raycheng, felixzhuo, harryyfhu, gavinzli}@tencent.com}

\renewcommand{\shortauthors}{Yuanzhen Xie et al.}

\begin{abstract}
% \red{Despite the ability of large language models to perform reasoning tasks by generating intermediate Chain of Thought reasoning steps, there remains a noticeable discrepancy between their capabilities and those of humans in solving complex reasoning problems. 
% It is widely recognized that humans typically employ various cognitive abilities when addressing complex reasoning challenges and necessitate interaction with tools, user feedback, and all aspects of the external environment to accomplish intricate tasks. 
% Consequently, we incorporate a human-cognition framework into complex reasoning to aid the model in comprehending sophisticated intentions and reducing problem-solving difficulties. 
% This paper presents a framework for intelligent agents that adheres to human cognition principles in combination with a large language model, which we denote as OlaGPT. 
% This model comprises several primary modules, including intention enhancement, memory, active learning, thinking, controller, and voting. The efficacy of OlaGPT has been stringently evaluated on multiple reasoning datasets, and the experimental outcomes reveal that OlaGPT surpasses state-of-the-art benchmarks, demonstrating its superior performance.}

In most current research, large language models (LLMs) are able to perform reasoning tasks by generating chains of thought through the guidance of specific prompts. However, there still exists a significant discrepancy between their capability in solving complex reasoning problems and that of humans.
At present, most approaches focus on chains of thought (COT) and tool use, without considering the adoption and application of human cognitive frameworks. It is well-known that when confronting complex reasoning challenges, humans typically employ various cognitive abilities, and necessitate interaction with all aspects of tools, knowledge, and the external environment information to accomplish intricate tasks.
This paper introduces a novel intelligent framework, referred to as OlaGPT. OlaGPT carefully studied a cognitive architecture framework, and propose to simulate certain aspects of human cognition. The framework involves approximating different cognitive modules, including attention, memory, reasoning, learning, and corresponding scheduling and decision-making mechanisms. Inspired by the active learning mechanism of human beings, it proposes a learning unit to record previous mistakes and expert opinions, and dynamically refer to them to strengthen their ability to solve similar problems. The paper also outlines common effective reasoning frameworks for human problem-solving and designs Chain-of-Thought (COT) templates accordingly. A comprehensive decision-making mechanism is also proposed to maximize model accuracy.
The efficacy of OlaGPT has been stringently evaluated on multiple reasoning datasets, and the experimental outcomes reveal that OlaGPT surpasses state-of-the-art benchmarks, demonstrating its superior performance.
Our implementation of OlaGPT is available on GitHub: \url{https://github.com/oladata-team/OlaGPT}.

\end{abstract}



% \keywords{human-like framework, intelligent agents, LLMs, Chain-of-Thought}


\received{20 February 2007}
\received[revised]{12 March 2009}
\received[accepted]{5 June 2009}

\maketitle

\section{Introduction}
In the past few years, large language models (LLMs) have developed the ability to process contextual information and generate fluent human language. As we encounter their outputs that sound natural and confident, we quickly assume that they have acquired the long-awaited thinking abilities, such as reasoning, communication, or collaboration, which are highly complex human skills. However, after in-depth understanding of LLM, we find that this reproduction based on high-probability language patterns is still far from the artificial general intelligence we expected. The most obvious gaps include the following: one is that LLMs in some cases produce content that is meaningless or deviates from human value preferences, or even dangerous suggestions with high confidence; secondly, the knowledge of LLMs is limited to the concepts and facts explicitly encountered in their training data. As a result, when faced with more complex problems, LLMs struggle to truly emulate human intelligence by understanding the ever-changing environment, collecting existing knowledge or tools, reflecting on historical lessons, decomposing problems, and using the thinking patterns that humans have summed up in the long-term evolution (such as Analogy, Inductive Reasoning and Deductive Reasoning, etc.) to effectively solve task. One way to solve the first problem is to introduce Reinforcement Learning from Human Feedback (RLHF), which has been recently implemented in ChatGPT \cite{brown2020language}. It attempts to explicitly encode human expression preferences into the training process: experts will be asked to rank the answers given by the model according to human common sense and ethical requirements. Obviously, this method is still an idea based on selection or injecting constraints, which can avoid the generation of toxic information to a certain extent, but still cannot reduce the gap with human reasoning ability.

Naturally, inspired by the attempts to use a combination of data weights and bias to mimic how neurons work, we hope to draw on the cognitive model of the human brain and the thinking models developed in the long-term evolution process more comprehensively, and design corresponding system components to endow these cognitive structures or thinking processes to LLM, so as to approximately align the reasoning processes of humans and LLMs, and expect LLMs to be able to solve complex problems more effectively. Some recent works have tried to partially solve some problems, such as using vector databases to store and retrieve external domain knowledge in real time, hoping to improve the memory of LLMs and the ability to capture real-time knowledge; other works such as langchain \footnote{https://python.langchain.com/en/latest/modules/agents/how\_to\_guides.html} and toolfomer \cite{schick2023toolformer} are designed to be able to leverage available tools, etc. However, the process of mimicking the human brain to deal with problems still faces many systematic challenges:

{\bf Challenge 1: How to systematically imitate and encode the main modules in the human cognitive framework, and at the same time schedule the modules according to the general human reasoning patterns in a realizable way.} As mentioned before, existing works \cite{yao2022react, schick2023toolformer} do not comprehensively attempt to align human and LLM reasoning pipelines.

 % \red{}
{\bf Challenge 2: How to motivate LLMs to perform active learning like humans, that is, learn and evolve from historical mistakes or expert solutions to difficult problems?} Encoding the corrected answers by retraining model might be feasible, but it is obviously costly and inflexible. The common in-context learning \cite{paranjape2023art, diao2023active, zhang2022automatic, brown2020language} is more to explain instructions or patterns in a few-shot way. The large model still lacks a human-like thinking framework for wrongly answered questions or historical lessons, such as "reflection-memorizing-reference-reasoning" mental model.

{\bf Challenge 3. How can a LLM flexibly be able to leverage the diverse thinking patterns that human beings have evolved so as to improve its reasoning performance? } It is hard to adapt to various problems by designing a fixed and general thinking model. Just like human beings usually choose different thinking methods flexibly when facing different types of problems, such as analogical reasoning, deductive reasoning and so on.

% \red{What we thought human-like AI should be like. }
% 方法描述初稿
% 为了解决上述提到C1，我们借助认知架构框架\ref{}设计了一套智能体方案，旨在借助人类思想解决复杂问题。方法包括几大模块：意图增强；记忆；主动学习；思维；控制者；投票模块。
% 此外，我们依据人类解决问题的思想设计了许多思维agent模板，指引模型以人类思维习惯来解决问题。
% 为解决C2，考虑到主动学习中，重新训练模型的方案成本过大。本文引入错题笔记的概念，将顽固错题进行人工修正或者按提示直到自我修正为止后积累到笔记池中，做题的时候动态查阅笔记找到相似题解法引入作为参考。
% 为解决C3，考虑到难拥有一套模板能兼容各种各样的问题，我们设计走多种思维模板之后再进行投票选择的方式。充分利用每种思维的优势之处，尽可能提升大模型回答的准确率。

In order to solve challenge 1, we carefully studied humans' cognitive architecture framework \cite{kotseruba202040}, designed some functions to approximate these thinking modules, such as understanding of intention, memory and comprehensive decision-making, etc., and designed the corresponding scheduling mechanism. To address Challenge 2, we propose a concept called "difficult question notes" with the aim of recording cases where the model frequently answers incorrectly. Notes for difficult cases are collected either through manual corrections or by following prompts until a self-correction is made. When answering questions, LLMs can dynamically review the note pool, identify solutions for similar problems, and use them as a point of reference. For the third challenge, we summarize the most effective reasoning frameworks for humans in problem solving, and design corresponding Chain-Of-Thought templates accordingly. We also designed a comprehensive decision-making mechanism to summarize the answers given by each COT reasoning, so as to maximize the accuracy of the model.
% 贡献
The main contributions of this work can be summarized as follows: 
\begin{itemize}
\item {} As far as we know, this is the first work that attempts to systematically enhance LLMs' problem-solving abilities by learning from a human cognitive processing framework. Results show that this alignment approach can boost the LLMs' performance in many aspects, such as understanding of intention, accuracy of knowledge, correctness of reasoning, etc.
% 据我们所知是首次结合大语言模型与人类认知架构设计仿人类思维的系统，极大提升了大语言模型的性能。
\item {} This work tries to summarize various methods of human reasoning into Chain-of-Thought (CoT) templates, so as to maximize the LLMs' reasoning effect in different scenarios. The article also innovatively designs an efficient active learning mechanism and vote mechanism to improve the accuracy and robustness of solving complex cases.
% 设计多种思维方式模板激发大模型的类人思维潜能，利用多思维模板投票的方式提升模型对不同题型的鲁棒性；错题笔记的构建和检索以让大模型实现新型主动学习方式。【最好给些案例比较，让别人看到思维引导下的结果比直接回答的要好且跟人类思维步骤相似】
\item {} We conduct comprehensive experiments on two datasets and evaluate each module of the proposed method. The experimental results demonstrate that OlaGPT outperforms state-of-the-art baselines, indicating its superior performance.

\end{itemize}

\begin{figure*}[ht]
  \centering
  \includegraphics[width=1.0\linewidth]{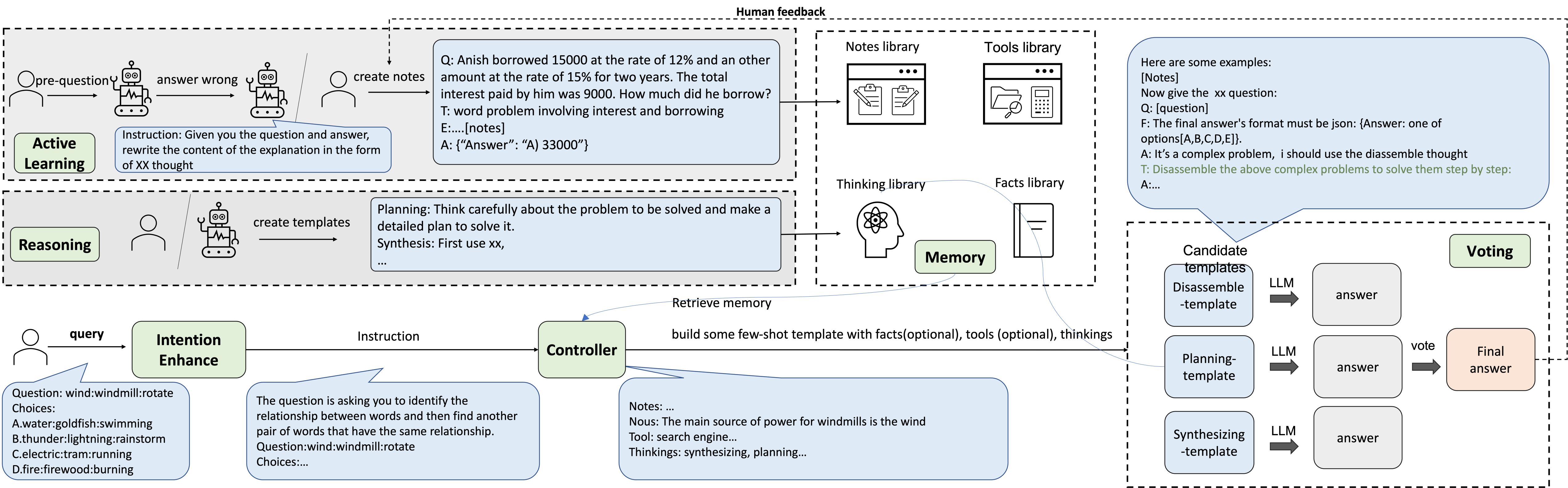}
  \caption{ The overall structure of the OlaGPT model.}
  \Description{A figure shows the overall structure of the OlaGPT model.}
  \label{fig:sys}
\end{figure*}

\section{Related Work}

\begin{table}[t]
 % \vspace{-0.67cm}
 \caption{Comparing OlaGPT with related approaches.}
 \label{data:main}
 \setlength{\tabcolsep}{1.0mm}
 \centering
 \small
 %\footnotesize
  \begin{tabular}{@{}c|ccccc@{}}
  \toprule
  \toprule
Features & CoT & Auto-CoT & Toolformer & OlaGPT \\
\hline
Multi-step reasoning & \checkmark & \checkmark &   & \checkmark\\
Limited supervision &  & \checkmark  & \checkmark & \checkmark  \\
Tool use &  &  & \checkmark & \checkmark \\
Extendable libraries &  &  &   & \checkmark \\
Cross-task transfer &  & \checkmark & \checkmark  & \checkmark\\
Human feedback & \checkmark &  &  & \checkmark\\
Active learning  &  &  &  & \checkmark \\

\midrule 

\bottomrule\hline
\end{tabular} \label{tab:com}\end{table}

\subsection{Augmented Language Model}
Augmented language model usually refers to the enhancement of a large language model's reasoning skills and the ability to use tools. The former is defined as decomposing a potentially complex task into simpler subtasks while the latter consists in calling external modules. 
% Prompt engineering refers to the method of using prompt techniques to guide its generation behavior without updating the weight of the model, so as to achieve the goal as much as possible. Relying on a large amount of human experience, the same prompt has large differences in different models.

\paragraph{\bf Chain of Thought}
Among prompt engineering methods, Chain-of-Thought (CoT) prompting \cite{wei2023chainofthought} is a popular technique that does not require fine-tuning model parameters. It is particularly effective in improving the model's performance in complex reasoning questions by simply changing the input. \cite{diao2023active} references uncertainty-based active learning to mine most uncertain questions, which are then manually annotated and iteratively selected to stimulate reasoning ability as much as possible. To reduce the cost of manual annotation, a fully automated pipeline named  Automate-CoT (Automatic Prompt Augmentation and Selection with Chain-of-Thought) is proposed in \cite{shum2023automatic}, which uses a variance-reduced policy gradient strategy to estimate the significance of each example in LLM. Another automatic CoT prompting method: Auto-CoT \cite{zhang2022automatic} samples questions with diversity and generates reasoning chains to construct prompt. Additionally, a solution is proposed in \cite{wang2023selfconsistency} to improve the results by using a voting strategy to select the most consistent answer output based on the results generated from different reasoning paths. Automatic Reasoning and Tool-use (ART) \cite{paranjape2023art} uses frozen LLMs to automatically generate intermediate reasoning steps as a program. This framework selects demonstrations of multi-step reasoning and tool use from a task library. Self-Taught Reasoner (STaR)\cite{zelikman2022star} relies on a simple loop: if the generated answers are wrong, try again to generate a rationale given the correct answer; fine-tune on all the rationales that ultimately yielded correct answers.
Few-shot prompting struggles as task complexity increases. Consequently, many recent works employ in-context learning to decompose complex problems into sub-problems and effectively teach these sub-problems via separate prompts, such as SeqZero\cite{yang2022seqzero}, Decomposed Prompting\cite{khot2023decomposed}.
\paragraph{\bf Tool Use}
Although recent LLMs are able to correctly decompose many problems, they are still prone to errors when dealing with performing complex arithmetics. Program-Aided Language models (PAL) \cite{gao2023pal} decompose symbolic reasoning, mathematical reasoning, or algorithmic tasks into intermediate steps along with python code for each step. Similarly, \cite{Drori_2022} prompts Codex\cite{chen2021evaluating} to generate executable code-based solutions to university-level problems. furthermore, \cite{schick2023toolformer} introduces Toolformer, a model trained to decide which APIs to call, when to call them, what arguments to pass, and how to best incorporate the results into future token prediction.

% 我们的工作从类人解决问题能力出发提升大模型性能，与部分相关工作的对比如表\ref{tab:com}所示。
Our work aims to enhance the performance of large models by drawing inspiration from human-like problem-solving capabilities, with a comparison to some related works presented in Table \ref{tab:com}.

\subsection{Cognitive Architecture}
Cognitive architecture is a subset of general artificial intelligence research that began in the 1950s with the ultimate goal of modeling the human mind, bringing us closer to building human-level artificial intelligence.
An early approach to cognitive architectures consisted of production systems, which used condition-action rules to represent and perform reasoning \cite{newell1972human}. 
One notable cognitive architecture resulting from this line of research is SOAR, which combined problem-solving, learning, and knowledge representation within a unified system \cite{laird1987soar}. SOAR has evolved into a comprehensive framework for modeling various cognitive processes such as decision-making, planning, and natural language understanding \cite{laird2019soar}.
Another influential cognitive architecture is ACT-R (Adaptive Control of Thought-Rational), which emphasizes symbolic processing and focuses on memory processes \cite{anderson2014atomic}. ACT-R has been applied to extensive cognitive tasks, such as problem-solving, language comprehension, and learning \cite{anderson2009can}.
DUAL (Distributed Unit for Assembling Learning) is a hybrid cognitive architecture that strives to balance symbolic processing and connectionist approaches \cite{kokinov1994hybrid}. By employing a central executive agent and multiple collateral units, DUAL manages a diverse range of cognitive tasks.
The Sigma architecture \cite{rosenbloom2016sigma} is composed of several key components, such as perception, working memory, long-term memory, production memory, subgoals, decision network, and learning mechanism. \cite{kotseruba202040} highlights the core cognitive abilities of these architectures and their practical applications in various domains. This paper will follow the framework of establishing an analogy between LLM and Cognitive Architectures.

\section{Method}
To address the challenges mentioned, we present a novel framework denoted as OlaGPT, which is a human-like problem-solving framework to empower LLMs.

% 
% \blue{To address the challenges mentioned, we present a novel framework denoted as OlaGPT, which is a human-like problem-solving framework based on LLMs.}

\subsection{Overview}

\begin{figure}[tbp]
  \centering
  \includegraphics[width=1.0\linewidth]{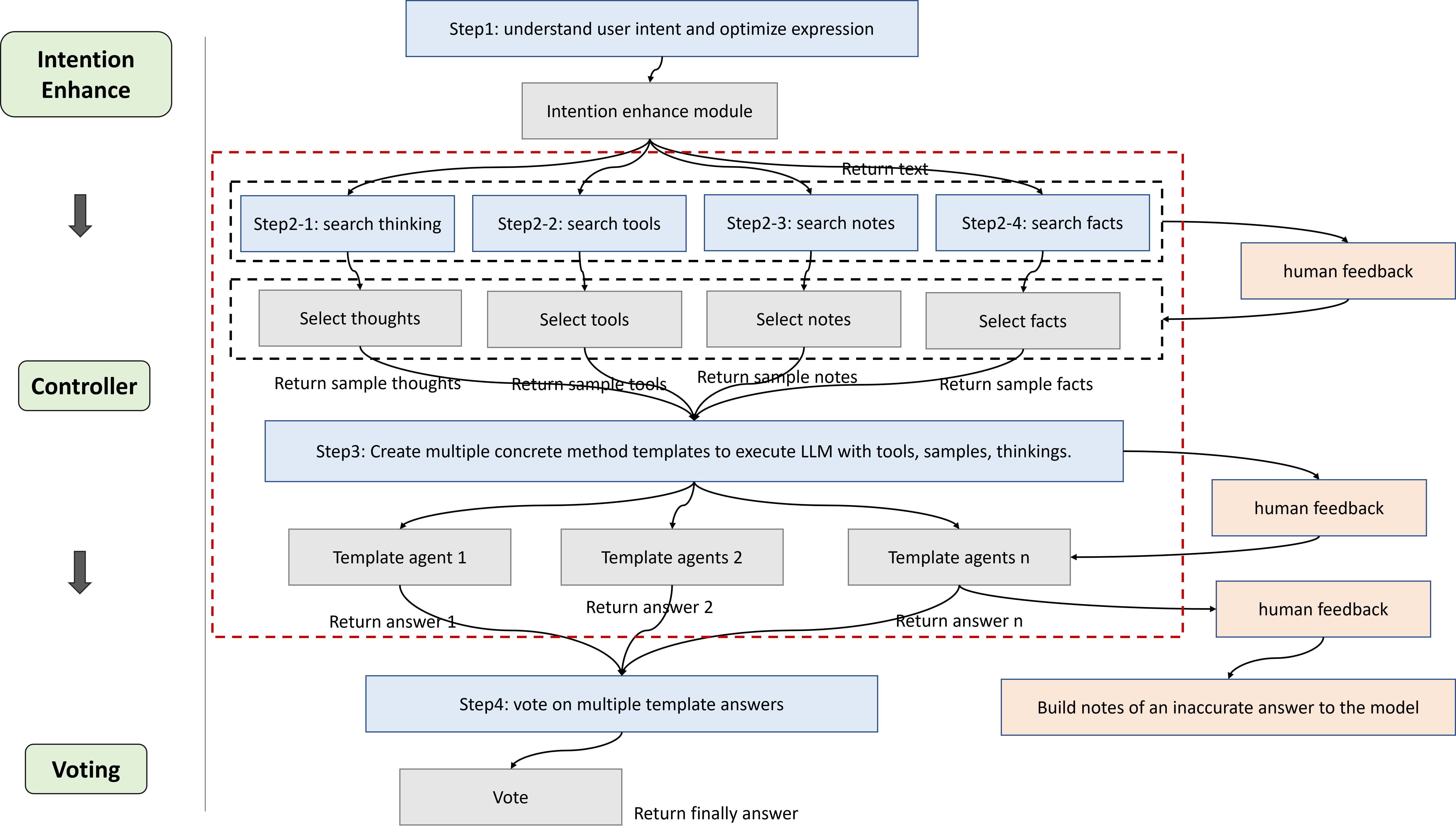}
  \caption{The overall flow chart of the OlaGPT model. First, OlaGPT enhances the user's question intention; the second step is to select multiple thinking templates, tools, wrong question notes, and factual knowledge that may be used; In the third step, the previously obtained wrong notes(used as examples), factual knowledge(pre-knowledge), thinking(guidance) are combined to complete the construction of multiple template agents, and then execute these agents to obtain preliminary answers; 
  The fourth step is to vote on the answers to get the final answer.}
  \Description{A figure shows the overall flow chart of the OlaGPT model.}
  \label{fig:flow}
\end{figure}

% To address the current limitations of LLMs and enhance their ability to mimic human thought, we have developed a human-like intelligent LLM solution based on research in human cognitive behavior. Our approach draws on the theory of cognitive architecture \cite{kotseruba202040} and comprises several core modules: the Intention Enhance module (corresponding to Attention), Controller module (Action Selection), Memory module (Memory), Active Learning module (Learning), Thinking module (Reasoning), and Voting module. We have summarized our intelligent simulator based on human cognition as OlaGPT, and begin with an outline of the methodology followed by a more detailed description of each component.
Our approach draws on the theory of cognitive architecture \cite{kotseruba202040}, which suggests that the core capabilities of the cognitive framework include {\bf Attention, Memory, Learning, Reasoning, Action Selection} \footnote{Relevant content is described in Appendix A.2.}. We fine-tuned this framework according to the needs of implementation and proposed a process suitable for LLM to solve complex problems. It includes six modules: {\bf the Intention Enhance module (corresponding to Attention), Memory module (Memory), Active Learning module (Learning), Reasoning module (Reasoning), Controller module (Action Selection), and Voting module}. 

% 增加角注？相关内容在附录A.2中阐述。

% 每个模块的功能简介如下：
The functional profile of each module is as follows:

\paragraph{\bf Intention Enhance.}
% The intent enhancement module focuses on enhancing the model's perception of user content, with the purpose of extracting the content that should be the most concern and better connecting user input and model expression habits.
% The Intention Enhance module focuses on enhancing the model's understanding of user queries, with the aim of extracting the most relevant information and better connecting user input with the model's expression habits.

According to \cite{kotseruba202040}, attention is an important component of human cognition, as it facilitates the identification of pertinent information and filters out irrelevant data. Similarly, we design corresponding attention modules for LLMs, namely Intention Enhance, aiming to extract the most relevant information and establish stronger associations between user input and the model's linguistic patterns.
% 本文中我们使用大模型和人类经验对提问进行优化。

\paragraph{\bf Memory.}
% 有研究表示大模型目前对最近的事实数据掌握能力相对匮乏，我们设计记忆模块重点将模型未能很好固化的现有知识，利用外置库的方式固定作为长期记忆。在问题询问过程中可以通过选择模块的检索功能将相关知识提取出来。
% Memory模块最重要的功能是存储一些库中的信息
% The Memory module stores information in some libraries
% The most important function of the Memory module is to store information in some libraries.
% Some studies \cite{schick2023toolformer, mialon2023augmented} have indicated that the LLMs currently lack the ability to grasp the latest factual data. 
% We design the Memory module to focus on the existing knowledge that the model has not solidified well and put them in external libraries as long-term memory. 
% % In the process of asking questions, relevant knowledge can be extracted by selecting the retrieval function of the module.
% There are four types of memory libraries involved in our paper: facts, tools, notes, and thought. See section 3.3 for details.

The Memory module serves a crucial function in storing information in various libraries. Recent studies \cite{schick2023toolformer, mialon2023augmented} have highlighted the limitations of current LLMs in comprehending the latest factual data. In light of this issue, we have designed the Memory module to focus on consolidating the existing knowledge that the model has not yet solidified and storing it in external libraries as long-term memory. During querying, the module's retrieval function can extract relevant knowledge from these libraries. There are four types of memory libraries involved in our paper: facts, tools, notes, and thinking.

\paragraph{\bf Learning.}
% 主动学习模块目的是让模型通过学习案例提升自己较差的能力。本文主要探讨学习错题笔记。错题笔记类似人类错题本的概念，主要意图将大模型目前无法解决的类型题找到并给予专家解题步骤和错因让它学会该类型题的处理。将错题沉淀成长期记忆库，在提问时检索与提问相关的错题引入一定的先验知识以让大模型更好地完成相关工作。
% 工具制作则意图在人类没有提供可用工具的时候，让大模型能够通过自己或者在人类协助下制作一些简易工具来完成工作。
% The purpose of the Active Learning module is to improve LLMs' poor ability through learning cases. 
% The paper mainly focuses on learning hard cases that LLMs can't answer well. First, we find some problems that cannot be solved by LLMs. Then, the experience and explanations written by experts are recorded in the notes library. Finally, the relevant notes are selected to LLMs to learn how to deal with these types of questions well when someone answers some similar question.

% Mistake notes are similar to the concept of human error books. 
% The main purpose is to find the types of problems that cannot be solved by LLMs and give experts the steps to solve the problems and the causes of mistakes so that they can learn how to deal with these types of problems. 
% 将经验累积记录到图书馆中
% Precipitate the wrong questions into a long-term memory library, and retrieve the notes related to the question when asking questions to introduce certain prior knowledge so that the large model can better complete the relevant work.

% 
The capability to learn is essential for humans to continuously improve their performance. Essentially, all forms of learning depend on experience. In particular, we found one way to quickly improve LLMs' reasoning ability is to let them learn from the mistakes it has made before. Firstly, we identify problems that cannot be resolved by LLMs. Next, we record the insights and explanations provided by experts in the notes library. Finally, we select relevant notes to facilitate LLMs' learning and enable them to handle similar questions more effectively.

\paragraph{\bf Reasoning.}
% 思考模块重点依据人的推理过程思想设计大模型的思维链能力，重点解决大模型的推理能力问题。具体设计上从思维技巧入手，设计了多种思维模板供大模型选择。
% 具体一个问题可以选择最适合的N种思维模板进行解答，最终再经过投票模块以提升准确率性能。[vote模块提及]
% Thinking module focuses on designing many agents based on the human reasoning process, stimulating the potential thought ability of LLMs to solve reasoning problems well.
% Various thought templates that reference thought skills, such as \textcolor{red}{analogy, decomposition, and plan}, are designed for specific reasoning tasks.

% Thinking module focuses on designing the thought chain ability of the large model based on the human reasoning process and focuses on solving the problem of the reasoning ability of the large model. 
% In terms of specific design, a variety of thought templates are designed for the large model to choose from. 
%  For a specific question, you can choose the most suitable N thought templates to answer, and finally go through the Voting module to improve the accuracy performance.

The Reasoning module is designed to create multiple agents based on the human reasoning process, thereby stimulating the potential thought capacity of LLMs to effectively solve reasoning problems. The module incorporates various thinking templates that reference specific thinking types, such as lateral, sequential, critical, and integrative thinking, to facilitate reasoning tasks.

% analogy, decomposition, and plan, to facilitate reasoning tasks.

\paragraph{\bf Controller.}
% 选择模块的目的是完成相关动作选择，对应论文\cite{kotseruba202040}中的Action selection。具体地，动作选择在本文中涉及到模型对任务的一些内在规划以及模版、工具、记忆的选择等。
% 是否选择某些模块
% the Controller module controls the use of some libraries and the selection of related content;
% The purpose of the Controller module is to control the relevant action selection, corresponding to the Action selection in the paper \cite{kotseruba202040}. Specifically, action selection in this paper involves internal planning of the model for tasks such as selecting certain modules which should be executed and choosing from libraries of facts, tools, notes, and thoughts libraries.

The Controller module is designed to handle relevant action selection, corresponding to the Action selection discussed in \cite{kotseruba202040}. Specifically, the action selection in this paper involves the internal planning of the model for tasks such as selecting certain modules to execute and choosing from libraries of facts, tools, notes, and thinking.

\paragraph{\bf Voting.}
% The Voting module allows for collective decision-making.
% Due to some cognition like different thought templates may be good at different types of questions and two heads are better than one, we design the Voting module to complete ensemble calibration among multiple thought templates. 
% % with a mode to execute multiple thought templates and then vote for the answer. 
% % It is responsible for scoring the results of multiple thought templates obtained above to complete the final output.
% % the most consistent
% It is responsible for generating the best answer by some voting strategies to improve performance.

The Voting module enables collective decision-making by leveraging the strengths of multiple thinking templates. As different thinking templates may be better suited for different types of questions, we have designed the Voting module to facilitate ensemble calibration among multiple thinking templates. The module is responsible for generating the best answer by employing various voting strategies to improve performance.

% [步骤投票]When disorientated majority processes overwhelm reasonable minority processes, the step-aware voting versifier can alleviate the limitation of vanilla majority vote
After introducing each module, we begin with an outline of the intelligent simulator with human-like problem-solving abilities (OlaGPT) followed by a more detailed description of each component.
% As illustrated in Figure \ref{fig:flow}, once the user inputs a query, the Intention Enhance module generates a more suitable question-and-answer format for the LLMs. The Controller module then retrieves potential tools, notes, factual knowledge, and multiple thought templates based on the user's intention. The required tools can be integrated into the retrieved thought templates, while notes and common knowledge are introduced as few-shot information.
% 
% 辅助信息
As depicted in Fig \ref{fig:flow}, once the user inputs a query, the Intention Enhance module generates a more understandable QA format for the LLMs. The Controller module then retrieves tools, notes, factual knowledge, and multiple thinking templates based on the user's intention. The relevant tools are integrated into the retrieved thinking templates, while notes and factual knowledge serve as supplementary information.
% 如图 \ref{fig:flow} 所示，一旦用户输入查询，Intention Enhance 模块就会为 LLM 生成更合适的问答格式。然后控制器模块根据用户的意图检索潜在的工具、笔记、事实知识和多种思维模板。所需的工具被集成到检索到的思维模板中，而笔记和常识作为小样本信息被引入。
% In the specific implementation of a single thought template, the tool library can also be dynamically indexed based on the thought content. (This function is implemented in the code, but because the data set currently used does not have a good tool to use, we did not use the tool in the data experiment.)
% \textcolor{green}{}
(In our implementation, we build an index for the tool library and dynamically retrieve the relevant tools for each template from the index. It is worth noting that this feature has been implemented in the code. However, due to the lack of suitable tools in the current datasets, we did not utilize this feature in the experiment.)
% (代码中实现此功能，但因目前使用的数据集没有较好的工具使用，在数据实验中我们没有使用工具。)

% 考虑到单一思维难以解决多种问题且为了确保模型提供更稳健的答案，每次选择多个模板同步执行，而不仅仅是单一的思维模板。[TODO: 丰满一下]
% Given the difficulty of solving multiple problems with a single thought and to ensure that the model provides more robust answers, multiple templates are selected for simultaneous execution at a time.
% % To ensure the model provides more robust answers, multiple templates are chosen to execute synchronously each time, not just a single thought template.
% % 用此种方案还可以解决不同类型题需要不同思维来解决的问题
% % 在单个思维模板的具体实现中，还可以根据思维内容对工具库进行动态索引。为了确保模型提供更稳健的答案，每次选择多个模板同步执行，而不仅仅是单一的思维模板。显，
% % we not only select a single thought template for execution each time, but also choose multiple templates for simultaneous execution.
% After obtaining answers from multiple single-thought templates, we employ various voting strategies in the Voting module to output the final answer.
Obviously, utilizing a singular cognitive framework is insufficient for addressing the diverse range of complex problems effectively. Thus, it is essential to adopt a variety of thinking template in order to derive more holistic and precise solutions, which is widely employed in model ensembles to reduce the variance in model outputs.
More specifically, we execute multiple templates simultaneously and then employ various voting strategies in the Voting module to get the final answer.

% 目前大模型性能上无法同时与多个领域专家抗争，不过我们可以将大模型和专家的能力进行结合。如图 \ref{fig:flow} 所示，可以将人类反馈纳入框架的各个方面，例如在模型中思维模板的实际实施过程中、在错误问题的产生过程中以及在控制器模块。通过允许人类专家协助模型，我们可以提高其性能（也可以禁用人类反馈模块以实现完全自动化的过程）。直观地，通过引入人类专家协助的方式，告知模型错误点或者无法判断的内容等进行指引可以极大提升LLMs的性能。在我们的设计中也可以禁用人类反馈模块以实现完全自动化的过程。
% Currently, LLMs are unable to compete with multiple domain experts simultaneously in terms of performance. However, we can combine the capabilities of large models and experts to achieve better results.
% Human feedback can be incorporated into various aspects of the framework as figure \ref{fig:flow} shows, such as adding human feedback tools during the actual implementation of the thought template, the production of the notes library, and the selection in the Controller module.
% Intuitively, incorporating human experts to provide guidance on error points or ambiguous content can significantly enhance the performance of LLMs in some complex problems.
Currently, LLMs have not yet reached a level of performance where they can surpass experts in various fields simultaneously. Nonetheless, by amalgamating the versatility of LLMs and the proficiency of specialists, it is possible to attain superior outcomes.
As depicted in Fig \ref{fig:flow}, human feedback can be integrated into multiple aspects of the proposed framework. These aspects may include incorporating human feedback tools during the implementation of the thinking template, curating the notes library, and making selections in the Controller module. Such an approach ensures that the overall performance is enhanced through the combined expertise of humans and the framework itself.
Intuitively, the inclusion of human experts to express clearly ambiguous content can notably improve the performance of LLMs when tackling intricate and multifaceted problems.

% 将人类反馈纳入框架可以帮助模型向人类专家学习并提高其解决复杂问题的能力。这种方法还可以增强模型模仿人类思维的能力，并提供更准确可靠的答案。
% Intuitively, incorporating human feedback into the framework can help the model learn from human experts and improve its ability to solve complex problems. 
% This approach can also enhance the model's ability to mimic human thought and provide more accurate and reliable answers.
% The human feedback sub-module can also be disabled in our design to achieve a fully automated process.
By leveraging the insights and expertise of human domain experts, LLMs can better understand the nuances of complex problems and generate more relevant and contextually appropriate responses.
It is worth noting that the human feedback sub-module can be disabled in our design to achieve end-to-end reasoning.
% 本文实验为节省实验人力开销，选择自动化流程，代码保留专家引入修正能力。
% \textcolor{green}{}
In this experiment, in order to reduce the cost of labor and resource, we choose the end-to-end approach for the experiment.

% 下文将详细介绍每个模块的具体内容
The specific content of each module is described in detail below.

\subsection{Intention Enhance Module}
% % 作为一种用户表达习惯到模型表达习惯的优化转换器。初期的实验中，我们针对不同的数据集设计了较为适配的意图增强语句。我们预先用大模型得到题目的类型（例如属于语义对比题），在提问开头添加“XX”语句。为了方便对结果进行解析处理，在题目最后加入“xx”语句。
Intention Enhance Module can be regarded as an optimized converter from user expression habits to model expression habits. A more suitable intent enhancement statement is designed for LLMs. Specifically, we get the type of questions by LLMs through specific prompts (see Table \ref{pro:qt} in Appendix) in advance and then restructure the way the question is asked. 
As Fig \ref{fig:notes_example} shows, the sentence——"Now give you the XX(question type produced by LLM model, the prompt see Table \ref{pro:qt} in appendix) question and choices:" is added at the beginning of the question. To facilitate the analysis and processing of the results, the sentence——"The answer must end with JSON format: {Answer: one of options[A,B,C,D,E]}.") is added at the end of the content.

Additionally, we are currently trying to build an automated intention enhance module, set up a seed dataset and related instructions, and call the GPT interface to generate a batch of training data. Using the open-source LLaMA model \cite{touvron2023llama} and Lora \cite{hu2021lora} technology, fine-tuning is performed using the data generated by the instruction. Intended to implement a module that automates user input enhancements. This module is still under development and experimentation, and relevant experiment results will be released in the future if there is an effect.

% We use the large model to get the type of the question in advance (for example, it belongs to the semantic comparison question), and add the "Now give you the XX(question type produced by LLM model, the prompt see Table \ref{pro:qt} in appendix) question and choices:" sentence at the beginning of the question. 
% In order to facilitate the analysis and processing of the results, the "The answer must end with json format: {Answer: one of options[A,B,C,D,E]}." statement is added at the end of the title.
% 经过大量测试发现，语言模型对字符的数量不敏感，对计算不敏感。

\subsection{Memory Module}
% 记忆模块主要进行相关知识和对话的存储。短期记忆我们利用langchain的memory功能，长期记忆则指外置的一些仓库（常识库、工具库、思维库）。本文重点的知识有三大类：事实库、工具库和笔记库（错题库），它们都通过LLMs建立文本索引。它们的简介如下：
The memory module mainly stores relevant knowledge and dialogues. We use the memory function provided by langchain for short-term memory, and long-term memory is implemented by a Faiss-based vector database \cite{johnson2019billion}. There are four main categories of knowledge in our approach: facts, tools, notes, and thinking library.
% all of which build text indexes using the Facebook AI Similarity Search (Faiss) method. 
% Their brief introductions are as follows:
We briefly describe these knowledge libraries as follows:

\paragraph{\bf Facts library}
Facts are real-world information like common sense and other knowledge that everyone accepts. 
%  for example, the content of Wikipedia belongs to this warehouse in our settings.
% 【todo：扩展下仓库的内容】
% 

\paragraph{\bf Tools library}
% 增强LLM几乎
% 为了解决大模型的一些缺陷问题，在不进行微调的情况下引入工具库来协助大模型完成相关工作。工具的形态只需要输入输出均为text即可,可以是搜索引起、计算器、维基百科库等，也可以是其他训练好的模型。
In order to solve some defects of LLMs, a tool library that contains search engines, calculators and Wikipedia libraries, is introduced to assist the LLMs to complete some work without fine-tuning. The input and output of the tool should be in text format.
%  which contains search engines, calculators, Wikipedia libraries, etc., or other trained models.
% 在工具层中我们目前尝试的有搜索引擎、计算器、维基百科库等。
% 经过大量测试发现，语言模型对字符的数量不敏感，对计算不敏感。
% 字符识别工具

\paragraph{\bf Notes library}
% notes主要记录大模型回答错或者是经典例题的解题思路（若为错题含错因），也可以成为经验。 区别于事实库，经验是对于过去环境（历史）的认知，并非完整反映客观历史。
% \textcolor{red}{Notes mainly record the wrong answer of the large model or the problem-solving ideas of the classic example (if the wrong question contains the cause of the error), it can also become experience. Different from the fact base, experience is the cognition of the past environment (history), not a complete reflection of objective history.
% }
% notes主要记录难的类型题以及 它们的解题步骤。
Notes mainly record some hard cases and their problem-solving steps. 
% 仓库的样例可查看表1
% todo: 构建可查阅3.4小节
Examples of Notes can be found in Fig \ref{fig:notes_example}.

\paragraph{\bf Thinking library}
% thoughts Library主要存放专家编写的人类解题思想模板。
Thinking library mainly stores human problem-solving thinking templates written by experts that can be humans or models.

% 举一反三

\subsection{Active Learning Module}

% 主动学习模块意图让模型主动学习一些知识，比如学习笔记内容、学习工具制作等。本小节主要研究笔记学习。
% The active learning module intends to let the model actively learn some knowledge, such as learning the content of notes, making learning tools, etc. This section mainly studies note-taking learning.
% LLMs存在一些自己的弱点题，Active Learning module致力于将它们找到并辅以专家知识进行解答。在需要回答相似题目时，可以将专家解答过程作为参考。
% Intuitively, LLMs have weaknesses in some specific tasks. The Active Learning module is dedicated to finding these hard cases and answering them by experts. LLMs can use the expert answering process as a reference when they meet similar tasks. In other words, the Active Learning module intends to let LLMs actively learn the types of questions that are often wrong. 

Intuitively, it is known that large language models (LLMs) have limitations in some specific tasks. The Active Learning module has been developed to identify challenging cases and provide expert answers. By doing so, LLMs can use expert answers as a reference when encountering similar tasks in the future. In essence, this module aims to facilitate the active learning process of LLMs, enabling them to acquire knowledge on the types of questions they typically struggle with.

% \red{In order to generalize problem-solving skills from a set of examples, we use notes as a reference answer.}

\paragraph{\bf Learn from mistakes}
% 特定场景下的语言模型的能力提升上常见的做法是对错题标注后进行重新训练，随着大语言模型的出现这种方法不太适用。因为重新训练大语言模型需要消耗非常巨大的资源，也无法做到实时更新。借鉴人类记录错题经验笔记的经验，本文给模型引入笔记库将顽固错题进行修正。当大模型做题的时候动态查阅错题集找到相似题解法引入作为参考。这种方式能通过prompt工程快速解决模型欠缺的部分解题能力。
% A common way to improve the ability of language models in specific scenarios is to retrain after labeling wrong questions. With the emergence of LLMs, this method is not suitable. 
% Because retraining a large language model consumes huge resources and cannot be updated in real-time. 
% Drawing on the experience of human beings in recording experience notes on wrong questions, the paper introduces a way of learning notes to improve the performance of LLMs. 
% First, we set up the notes library that LLMs always have made mistakes. Second, when the user asks a question, we can use the Controller module to select similar questions from the notes library dynamically as a guide. 
% % When working on a large model, dynamically consult the set of wrong questions to find solutions to similar questions and introduce them as a reference. 
% This way can quickly solve some problem-solving abilities that the model lacks through the prompt project. The specific question format can be seen in figure \ref{fig:notes_example}. 

\begin{figure}[tbp]
  \centering
  \includegraphics[width=1.0\linewidth]{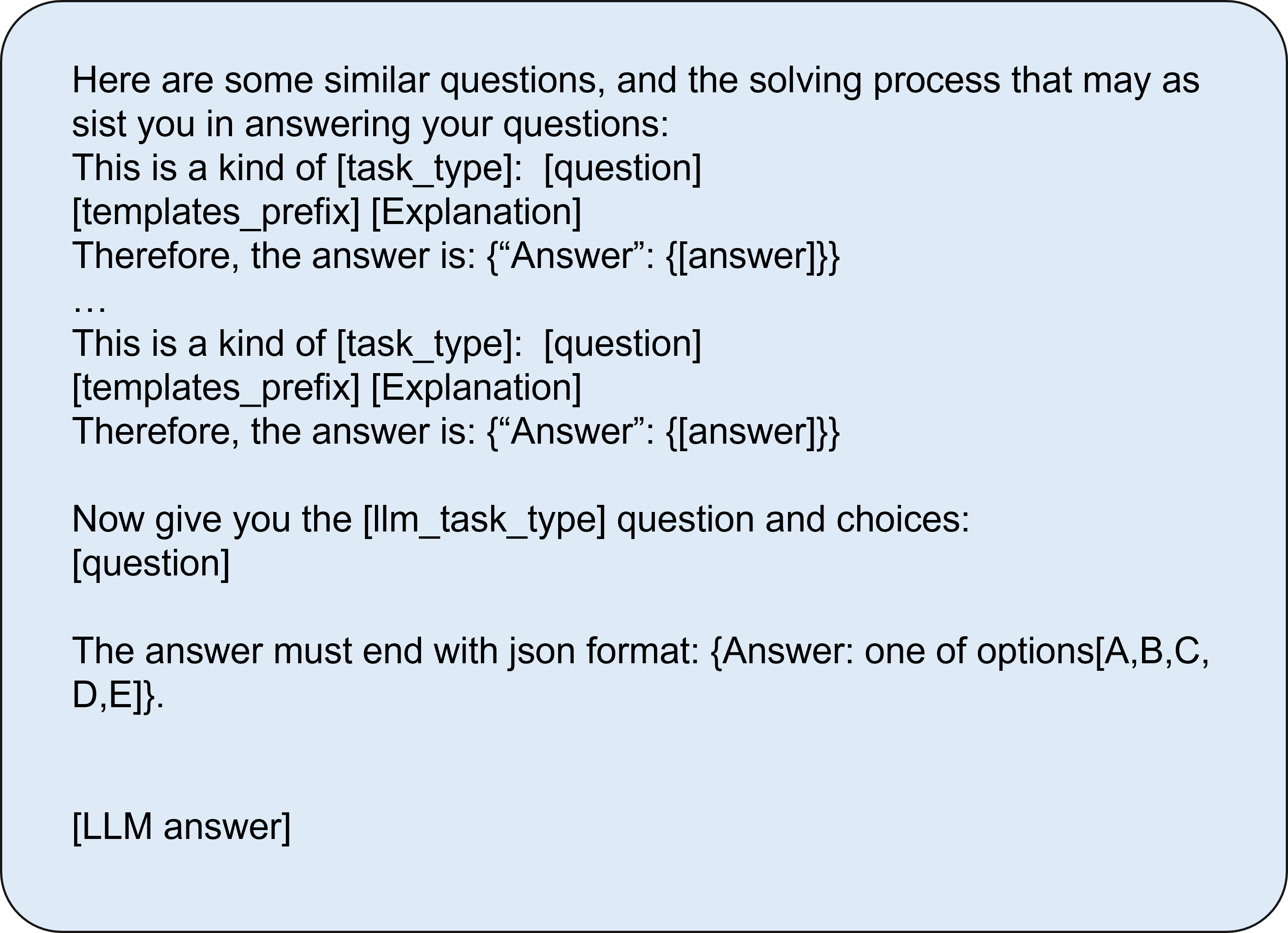}
  \caption{The notes are combined as examples. The templates\_prefix is template-specific content.}
  \label{fig:notes_example}
\end{figure}

\begin{figure*}[ht]
  \centering
  \includegraphics[width=1.0\linewidth]{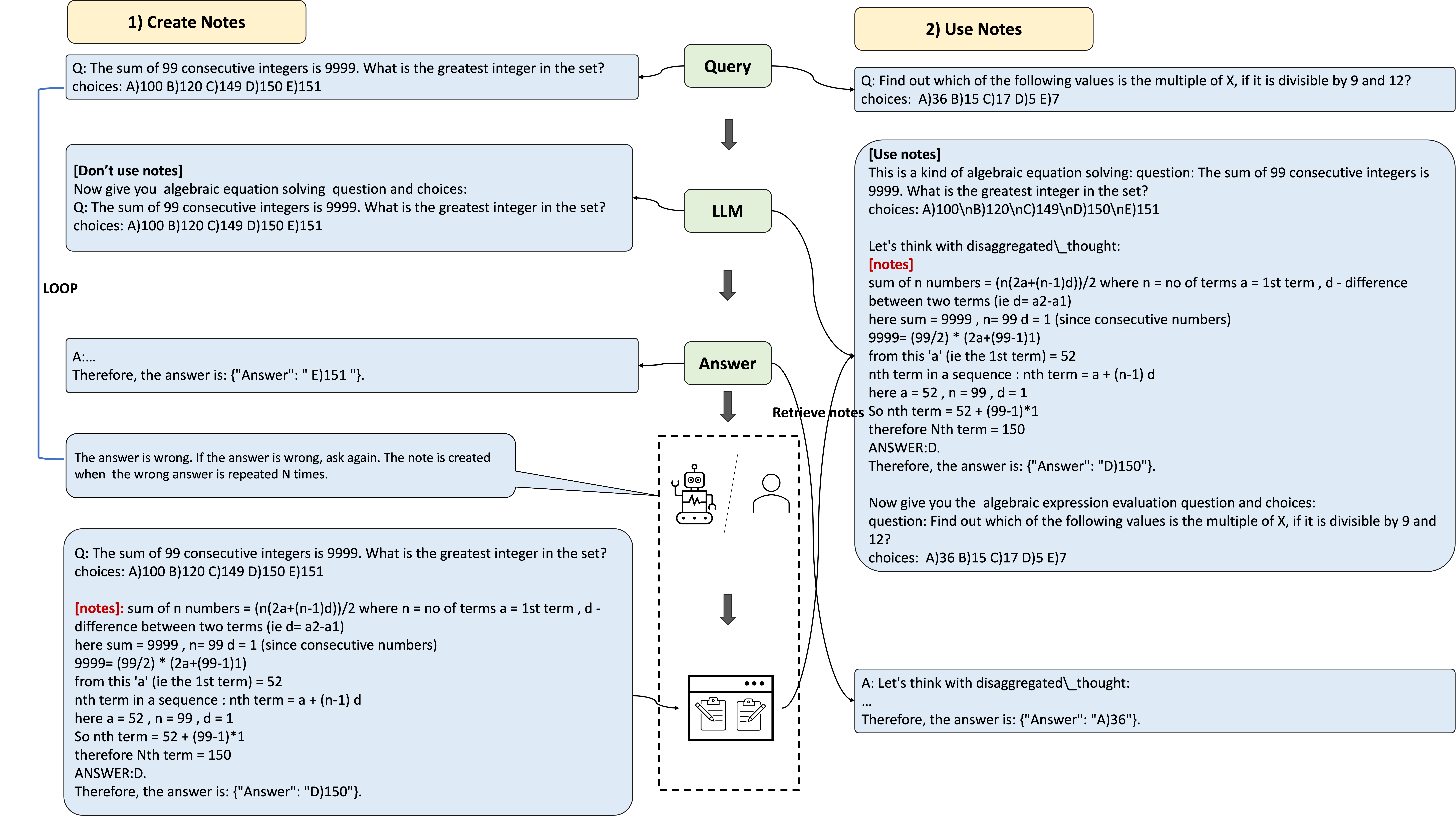}
  \caption{ The overall of the Active Learning module.}
  % \Description{A figure shows the overall structure of the OlaGPT model.}
  \label{fig:active}
\end{figure*}

A common approach to improving the performance of language models in specific scenarios is to retrain them with annotated answers\cite{radford2018improving}. However, with the emergence of LLMs, this method is not practical as it requires enormous resources and cannot be updated in real-time. Drawing inspiration from how humans use notes to record their mistakes, we propose to introduce the note library for models to correct stubborn mistakes. When the LLM encounters a difficult question, it can dynamically refer to the note library to find similar problems and their solutions as references as shown in Fig \ref{fig:active}. This approach can quickly improve the LLM's problem-solving abilities through prompts engineering. The specific notes format can be seen in Fig \ref{fig:notes_example}.
% 笔记学习想让模型主动学会常做错的类型题。在前文我们设置了模型曾经做错的笔记库知识，用户发起提问时候，我们会通过检索模块从笔记库中选择相似的题目类型作为引导。设置格式让模型从前文的题型中获取经验知识，具体提问格式如图所示：
% As described earlier, we set up the notes library that LLMs always have made mistakes. 
% When the user asks a question, we can use the Controller module to select similar questions from the notes library as a guide. The specific question format is shown in figure \ref{fig:notes_example}. 
% Set the format to allow the model to obtain empirical knowledge from the previous question types. 

% 【todo: 具体的图文解析】

\paragraph{\bf The creation of of notes}
% 笔记该如何构建呢？笔记的构建过程为首先让大模型重复回答一批问题，然后记录模型总是回答错误的问题，最后利用专家为它们添加解析和题目类型等，形成笔记图书馆。
% How should notes be structured? First, let LLMs answer a batch of questions repeatedly, then record the questions that LLMs always answer incorrectly, and finally let experts write the question types, detailed problem-solving steps, and general problem-solving ideas to form the notes library.
% These contents can be written by humans or can be answers to given questions and prompts to guide the output of LLMs. Intuitively, it would be better if it is written by an expert-level human being and then rewritten by a model.
When constructing notes library, it is recommended to first have LLMs answer a batch of questions repeatedly as shown in Fig \ref{fig:active}. The questions that LLMs consistently answer incorrectly can then be recorded. Subsequently, experts can ensure the question types, detailed problem-solving steps, and general problem-solving ideas to form the notes library. Although the notes can be written by either experts or LLMs, it is generally preferable to have experts write the notes first and then have LLMs refine them.
The final dataset format is processed as json: \{"question":"x", "answer":"x", "error\_reason":"x", "model\_expert":"x", "explanation":"x", llm\_task\_type :"x"\}.
The type of task is generated with LLMs by specific prompt as shown in Table \ref{pro:qt}.

\paragraph{\bf The strategies of retrieving notes}
In the Controller module, there are various ways to implement note retrieval. One intuitive approach is to find the most similar question type as an example reference, which has been found to be highly beneficial for LLMs. 
% However, the number of questions with the same type is often too large to use as a few-shot input. 
However, the number of questions of the same type is often too large to be used as additional knowledge input to LLMs (exceeding the number of tokens).
Thus we have implemented multiple strategies, including \emph{Random Selection}, \emph{Dual Retrieval}, and \emph{Combine}, denoting as $random$, $retrieval$ and $combine$ respectively. $random$ indicates the method that randomly selects questions from the notes library. $retrieval$ represents a dual retrieval strategy that first retrieves the most similar question type using LLM and then retrieves the top-n relevant notes from the notes of such question type. However, text similarity does not always indicate question similarity. Hence, for $combine$, we use a random retrieval method to increase diversity in the second retrieval stage of  $retrieval$. Last but not least, we use zero-shot to identify the zero-shot strategy, which means disabling the Active Learning module from the framework.

% 
% \blue{In the Controller module, there are various ways to implement note retrieval. One intuitive approach is to find the most similar question type as a few-shot, which has been found to be highly beneficial for LLMs. However, the number of questions with the same type is often too large to use as a few-shot input. Thus we have implemented multiple strategies, including \emph{Random Selection}, \emph{Dual Retrieval}, and \emph{Combine}, denoting as $few\_shot=1$, $few\_shot=2$ and $few\_shot=3$ respectively. $few\_shot=1$ indicates the method that randomly selects questions from the notes library. $few\_shot=2$ represents a dual retrieval strategy that first retrieves the most similar question type using LLM and then retrieves the top-n relevant notes from the notes of such question type. However, text similarity does not always indicate question similarity. Hence, for $few\_shot=3$, we use a random retrieval method to increase diversity in the second retrieval stage of $few\_shot=2$. Last but not least, we use $few\_shot=0$ to identify the zero-shot strategy, which means disabling the active learning module from the framework.}

\subsection{Reasoning Module}
\begin{figure*}[ht]
  \centering
  \includegraphics[width=1.0\linewidth]{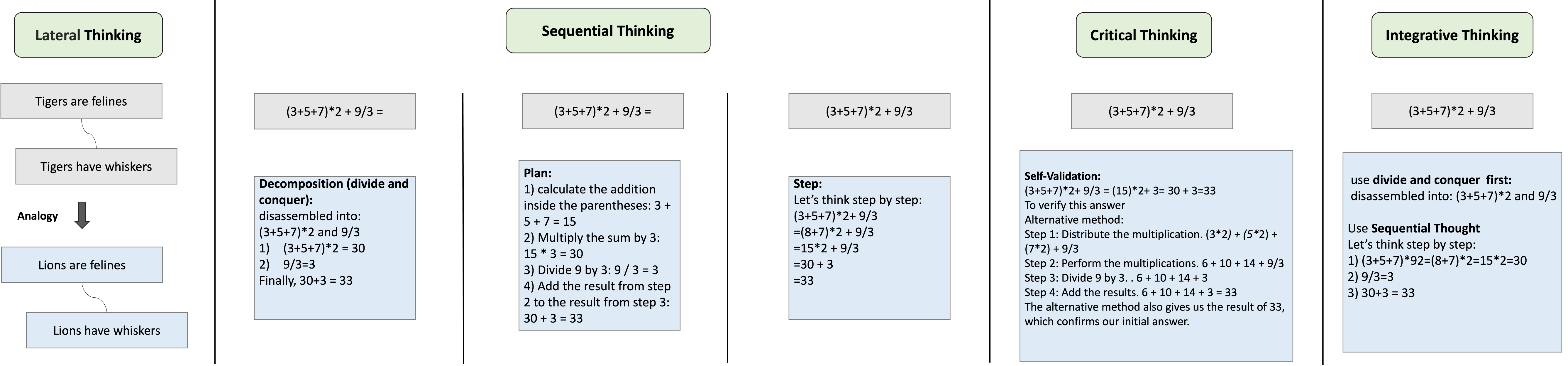}
  \caption{ The example of some thinking.}
  % \Description{A figure shows the overall structure of the OlaGPT model.}
  \label{fig:thinking}
\end{figure*}

% 人类具有的各种思想是支撑解题正确、做事完善的一大核心关键。为尽可能挖掘模型的推理能力，我们在思想模块设计了多种人类推理思路来帮助模型进行推理。
% 按照思维技巧的方式进行设计，思维技巧包括反思、验证、分解、归纳（类比）、综合、对比和规划思维等。
% 验证思维的主要思想是使用 LLM 生成输出，然后允许同一模型为其自身的输出提供多方面的反馈；最后，同一模型根据自己的反馈改进其先前生成的输出。分解思维则引将复杂问题拆解为子问题，主要引导做任务分解。归纳思维意图从一些样例中归纳总结解题技巧，实验中是从few-shot里面作为参考解答。综合思维是用多种方式结合。具体实现中引导模型使用多种思想，类似自由DIY的模式，该思想可以自由选择其他已有思想进行自由组合，以更好地完成问题的解答。
% % 本文在实验中探索了decomposition, comparison, planning, 下面主要介绍下实验中使用到的思维模板。
The diverse cognitive processes that humans possess are of importance for accurate problem-solving and achieving excellence in various tasks. 
To fully leverage the model's reasoning ability, we have designed multiple human reasoning approaches in the Reasoning module to assist the model in reasoning. 
% These approaches were designed based on thought skills, including reflection, verification, decomposition, induction, analogy, synthesis, comparison, planning thinking, and so on. 
These approaches are designed based on four thinking types, including Analogical, Sequential, Critical, and Synthesis thinking, which show in Fig \ref{fig:thinking}. The following introduces some thinking principles and implementation methods.

% [todo: 找个参考论文]
%  and we have selected several commonly used ones, including verification, decomposition, induction (analogy), synthesis, comparison, and planning.

\begin{itemize}
% [leftmargin=*]
    \item {\textbf{Lateral Thinking }}: Lateral Thinking is a creative problem-solving approach that involves looking at situations from unconventional perspectives. It includes techniques such as challenging assumptions, seeking alternative solutions with analogy, and embracing ambiguity.
        \begin{itemize}
            \item {\textbf{Analogical Thinking}}: Analogy Thinking is a cognitive process that involves finding similarities between two distinct entities or concepts in order to gain a deeper understanding of a problem or situation, which is one of the most effective tools to generate innovative ideas \cite{kim2016analogical}. The prompts can be seen in Table \ref{pro:tem}.
        \end{itemize}
    % 类比思想
    % Analogical Thinking for Generation of Innovative Ideas
    % https://medium.com/the-creative-mind/analogical-thinking-a-way-to-produce-creative-ideas-510cb9923fd0
    
    \item {\textbf{Sequential Thinking}}: Sequential Thinking (also named Linear Thinking) is a problem-solving approach that involves breaking down complex tasks or problems into smaller, more manageable parts and addressing them in a logical, step-by-step manner \cite{groves2008linking}.
        \begin{itemize}
            \item {\textbf{Decomposition Thinking}}: Decomposition Thinking involves breaking down complex problems into smaller sub-problems and guiding task decomposition. 
            % Disassemble Thought
            \item {\textbf{Plan Thinking}}: Plan Thinking involves creating a detailed roadmap or strategy for achieving a specific goal or completing a task.
            % autogpt
            % 在遇到复杂问题时候适合
            %  It helps to increase efficiency and reduce errors by providing a clear direction and structure for decision-making and action-taking.
            \item {\textbf{Step Thinking}}: Step Thinking is a problem-solving approach that involves breaking down complex tasks into smaller, manageable steps, allowing for more efficient and organized solutions. 
        \end{itemize}
    
    % \cite{kojima2022large}
    %  It helps to reduce errors and increase efficiency by providing a clear roadmap for completing a task or achieving a goal. The prompts see in 
    % 分步骤完成的思维通常被称为“分步思考”或“系统思考”。这种思维方式强调将一个大问题或任务分解成更小的、可管理的部分，然后逐步解决每个部分，最终达到整体解决问题的目标。这种思维方式可以帮助人们更好地组织和管理复杂的任务，减少错误和失误的风险，并提高效率和准确性。
    \item {\textbf{Critical Thinking}}: 
    Critical thinking is the process of objectively analyzing and evaluating information to form reasoned judgments. 
    % It involves questioning assumptions, identifying biases, and seeking diverse perspectives to make well-informed decisions.
    It includes the component skills of analyzing arguments, making inferences using inductive or deductive reasoning, judging or evaluating, and making decisions or solving problems \cite{lai2011critical}.
        \begin{itemize}
            \item {\textbf{Verification Thinking}}:  In real-life problem-solving scenarios, humans often employ one approach to solve a problem and then use another approach or seek the input of another person to verify the solution and improve its accuracy. This collaborative and iterative process enables more reliable and well-rounded outcomes.
            % Self-validation theory: An integrative framework for understanding when thoughts become consequential
            % 据此我们设计了验证思维，
            % Based on this, we designed the verification thought, which is to use an LLM to generate an output and then allow the same model to feedback on its own output from multiple perspectives. The model can then improve its previously generated output based on this feedback. It has not been used in experiments so far.
            Based on this observation, we designed the verification thinking, which involves using an LLM to generate an output and then allowing the same model to provide feedback on its own output from multiple perspectives. The model can subsequently refine its previously generated output based on this feedback. It has not been used in experiments so far.
            % The main idea behind verification thought is to use an LLM to generate an output and then allow the same model to feedback on its own output from multiple perspectives. The model can then improve its previously generated output based on this feedback. 
            
        \end{itemize}
    
    \item {\textbf{Integrative Thinking}}: Integrative Thinking involves combining multiple approaches in various ways, which seems to be the core component in models of adult cognitive development \cite{kallio2011integrative}. In our specific implementation, we guide the model to use a variety of thought approaches in a free DIY(Do It Yourself) mode. As of yet, it has not been utilized in any experiments.
    % reflection
    
    % \item {\textbf{planning}}: Planning thought is to make a detailed plan first, and then complete one thing according to the plan.
    % 计划思维是先列好详实的计划，然后依据计划来完成一件事情。
\end{itemize}

Collectively, these thinking methods are designed to enable models to reason more efficiently and provide more accurate answers.
% 上述介绍的思维并不全，常见的还有Concrete Thinking， 抽象思维、Vertical Thinking等。实际使用时候可以依据情景设计不同类型的思维模版。
% 该模块中可以根据不同的任务增加对应的思维模板，也可以给定多种思维模板交由大模型自己组合选择。
% In this module, corresponding thinking templates can be added according to different tasks, or new templates can be combined and selected by LLMs with a variety of thinking templates.
The thinking introduced above is not exhaustive, and the common ones include Concrete Thinking, Abstract Thinking, Vertical Thinking, etc. In actual use, different types of thinking templates can be designed according to the situation.
In this module, appropriate thinking templates can be incorporated based on the specific requirements of different tasks. Alternatively, LLMs can generate new templates by combining and selecting from an array of existing thinking templates. This flexibility allows the model to adapt and engage more effectively with the diverse challenges it encounters.
In the experiment, the paper explores analogy (Lateral Thinking), decomposition (Sequential Thinking), plan (Sequential Thinking), and step (Sequential Thinking) thinking. 
% 

% 选择多个思维模板的原因有：1、不同的模板适用于不同的子类型题。2、多模板回答让结果更加鲁棒

% 【待定】对比：将错题引入进行题目的对比
% 规划思维（引导做计划——适用于需要多步骤前后完成的任务）
% 【待定】直觉思维（直接拿到的东西？？？

\subsection{Controller Module}
% 【todo】动态索引操作实现简介：

% 【todo：如何支持内容扩展的问题】
% As the second step shown in Figure \ref{fig:flow}, the Controller module is mainly responsible for the retrieval and matching of relevant facts, tools, notes, and thoughts. 
% Controller模块中先进行相关事实、工具、笔记、想法的检索和匹配（如图\ref{fig:flow}的第二步骤）。然后将检索得到所需的内容组合为思维智能体，最后通过异步的方式，请求大模型得到单模板下的答案（如图所示的第三步）
% In the Controller module, the relevant facts, tools, notes, and thoughts are retrieved and matched first (as shown in the second step of Figure \ref{fig:flow}). 
% Then combine the retrieved content into a thought agent, and finally request LLMs to get the answer under the single template in an asynchronous manner (as shown in the third step of Figure \ref{fig:flow}).
% It is unrealistic to expect users to select everything they need at the beginning of the reasoning process, just as it is difficult for humans to find everything they need at the beginning of reasoning. Thus dynamic retrieval is achieved based on the user's question and intermediate reasoning process. 
In the Controller module, relevant facts, tools, notes, and thinking are first retrieved and matched (as shown in the second step of Fig \ref{fig:flow}). The retrieved content is subsequently integrated into a template agent, requiring the LLMs to furnish a response under a single template in an asynchronous manner (as shown in the third step of Fig \ref{fig:flow}). Just as humans may struggle to identify all relevant information at the outset of reasoning, it is difficult to expect LLMs to do so. Therefore, dynamic retrieval is implemented based on the user's question and intermediate reasoning progress.
% % 前面提到已经使用 Faiss 方法为所有四个库建立嵌入索引，对他们的检索操作上有所不同。
% As mentioned earlier, the Faiss method has been used to build embedded indexes for all four libraries, and the retrieval operation for them is different.
% % 具体地，常识检索作为一种工具存在，在LLM模型的问题分解和推理过程中随时从外部知识库中读取。在工具选择层面，我们提供了工具列表及其功能描述，让大模型在推理过程中实时选择。notes的检索策略与其他的不同，我们设计了3.4节中描述的三种方式。关于thoughts的检索可以通过prompts的方式让大模型自己判断选择, 在实验中目前全选进行分析尝试。
% specifically, at the tool selection level, we provide a list of tools and their functional descriptions, allowing LLMs to choose in real-time during the reasoning process. 
% The facts retrieval exists as a tool and can be read at any time during the LLM model's problem decomposition and reasoning process. 
% The notes selection strategy is different from others, and we design three ways that are described in section 3.4.
% Regarding the retrieval of thoughts, we design prompts to allow the LLM to judge and choose by itself.

As mentioned earlier, the Faiss method \cite{johnson2019billion} has been employed to create embedding indices for all four libraries, with retrieval strategies differing among them. Specifically, information retrieval serves as a tool that can be accessed at any time during the LLM model's problem decomposition and reasoning process from external knowledge bases. In terms of tool selection, we provide a list of tools, enabling the LLM to make real-time choices during the reasoning process. The retrieval strategy for notes differs from the others, as we have designed the three methods described in Section 3.4. As for the retrieval of thinking, we design prompts to allow the LLM for judging and selecting relevant thinking templates. But currently, we use all thinking templates together to perform a comprehensive analysis in the experiment.

\subsection{Voting Module}
% 如图\ref{fig:flow}所示的第三步，我们通过LLMs获得多个思维模板的结果。直观上，不同的思维模板可能适用于不同类型的问题。
% 因此，在使用多个单一模板通过LLMs获得答案后，我们采取对多个答案进行投票的方法来提高最终性能。
% After the third step shown in Figure \ref{fig:flow}, We obtain the results of multiple thought templates through LLMs. Intuitively, different thought templates may be suitable for different types of questions. 
% Therefore, after using multiple single templates to obtain answers through LLMs, we take some methods of voting on multiple answers to improve the final performance and enhance the robustness of the model's results. 
Following the third step depicted in Fig \ref{fig:flow}, we acquire the answers of various thinking templates through LLMs. Intuitively, distinct thinking templates may be more suitable for different types of questions. Therefore, instead of relying on a single template, we adopt a voting mechanism to aggregate the answers from multiple templates and improve the final performance of our model. This approach also enhances the robustness of the model's results, as it takes into consideration the variability in the efficacy of different templates when applied to distinct questions.

% The selection of ideas by a model is dependent on the description of the given ideas, and there may be a knowledge gap that needs to be bridged. 
% In real-life problem-solving scenarios, humans often employ one approach to solve a problem and then use another approach or seek the input of another person to verify the solution and improve its accuracy. 
%  Thus, we voted on the execution results of multiple thought templates selected and assembled above. 

% Thus, to enhance the robustness of the model's results, we adopt some voting strategies where the results obtained from multiple approaches are considered and the most popular answer is selected.
% 从实验结果看，正则的方法更好。
% From the experimental results, the regular method is better.
% 我们将对多个思想模板执行的结果进行最终的投票。具体，我们将上述经过选择模块选出的错题笔记、相关工具等填充到选中的多个思维模板中。利用多个思维模板对用户的query进行异步解答，获取答案之后再经过vote模块。投票方式有以下几种：

% There are several voting methods available:
% 1) LLM model vote: Let the LLM model select the most consistent answer among several answers through prompts.
% 2) Regular vote: Let the large model be output according to a specific number, and analyze it through regular expression matching to see the result of the majority vote.

There are several voting methods available:
1) vote by LLM: Instruct the LLM model to choose the most consistent answer among several given options by providing an output answer along with a rationale through prompts.
2) vote by regex: Extract the answer by regex expression to get the majority vote.

\section{Experimental}
We conducted a series of experiments on a range of public datasets and compared the proposed OlaGPT with existing approaches. Our findings indicate that our approach consistently outperforms the baselines on every dataset considered, demonstrating its robustness and effectiveness.

\subsection{Experimental Setup}
Our experiments are designed to address the followings:
\begin{itemize} 
% [leftmargin=*]
    \item {\textbf{RQ1.}} How does OlaGPT perform compared to the state-of-the-art baselines?
    \item {\textbf{RQ2.}} % 思维模板的适配性
    How effective are sub-modules in the overall model design?
    \item {\textbf{RQ3.}} % 超参数对实验结果有何影响
    How do hyperparameters affect experimental results?

\end{itemize}

\paragraph{ \bf DataSets.}

\begin{table}[t]
 % \vspace{-0.67cm}
 \caption{Statistics of Datasets.}
 \label{data:stas}
 \setlength{\tabcolsep}{1.0mm}
 \centering
 \small
 %\footnotesize
  \begin{tabular}{@{}c|ccccc@{}}
  \toprule
Domain & Datasets & \#testing & \#training & \#error books \\
\midrule
\multirow{1}{*}{mathematical reasoning} & AQuA & 254 & 97467 & 81 \\
 % & GSM8K &  &  & \\
\midrule
\multirow{1}{*}{analogical reasoning} & E-KAR & 335 & 1155 & 632 \\

% & E-KAR-English & 262 & 868 & 501 \\

\midrule

\bottomrule\hline
\end{tabular} \label{tab1-1}\end{table}

% 为了更有效地评估提出的方案，我们使用了几个公开数据集进行实验。
% In order to evaluate the proposed scheme more effectively, we use several publicly available datasets for experiments. 
% \textbf{AQUA (Algebra QA with Rationales)}, a dataset of roughly 100,000 algebraic word problems and 254 test questions, which is sometimes referred to as the MATHQA dataset due to a follow-up work (Ling et al., 2017; Amini et al., 2019). We randomly sample 200 training problems for our labeled training set.
To evaluate the proposed framework in a more comprehensive manner, we utilize several publicly available datasets for experimentation. \textbf{AQuA} \cite{ling2017program} (Algebra QA with Rationales) is a dataset comprising approximately 100,000 algebraic word problems and 254 test questions. For our labeled training set, we randomly sample 200 training problems.
% A follow-up work by Ling et al. (2017) and Amini et al. (2019) also refer to this dataset as MATHQA.
% TODO: 这里的引用需要加上
% 分了20个类；从中按权重抽取211题。
% E-KAR是首个可解释的知识密集型类比推理数据集，由 1,655 个（中文）和 1,251 个（英文）来自中国公务员考试的问题组成。
\textbf{E-KAR} \cite{chen-etal-2022-e} is the first interpretable knowledge-intensive analogical reasoning dataset consisting of 1,655 (Chinese) and 1,251 (English) questions from the Chinese Civil Service Exam. 
% The Chinese one is used in our experiment.
% 这里原因是否需要补充，会被问为什么不使用 english version.
In our experiment, we utilize the Chinese version to explore the performance with  the Chinese language. 

% 具体，针对较多数据的类型，使用Sentence-bert {reimers2019sentence} 进行embeddding，然后做K-means聚类后挑选部分数据对大模型进行轮询。【放入数据处理中】
% 数据集统计结果如表\ref{data:stas}所示。将训练集中大模型3-5次选择均错误的作为大模型的错题集；因aqua原始训练集较为庞大，对训练集做了聚类后依据权重重新选择训练集：具体分了20个类，最终从中按权重抽取211题。
% The statistical results of the data set are shown in table \ref{data:stas}. The large model in the training set is selected incorrectly for 3-5 times as the wrong question set of the large model; because the original training set of aqua is relatively large, the training set is reselected according to the weight after clustering the training set: specifically divided into 20 categories, and finally extract 211 questions according to the weight.
The statistical results of the dataset are presented in Table \ref{data:stas}. We identified the questions that the large model answered incorrectly 3-5 times from the training set as its error books. However, it is too expensive to achieve this on the training set of \textbf{AQuA} due to its large size. To address this issue, we utilized Sentence-BERT \cite{reimers2019sentence} for embedding and then clustered the training set into 20 groups using K-means. Finally, 211 questions were randomly selected from each group based on their weights.
% TODO: 加入sentence bert的ref

\paragraph{\bf{Evaluation Metrics}}
% 测量指标为准确率，目前题目类型大多为选择题，与答案匹配则正确。匹配中利用正则解析和人工辅助校验的方式查阅模型的正确率。
% The measurement index is the accuracy rate. Most of the current question types are multiple-choice questions, and if they match the answers, they are correct. In the matching, the correct rate of the model is checked by means of regular analysis and manual assisted verification.
% The current measurement metric is accuracy rate, and the majority of the questions are of the multiple-choice type, where a match with the answer key is considered correct. To determine the model's accuracy rate, we employ a combination of regular expression analysis and manual verification.
The current metric for measurement is the accuracy rate, and most of the questions are in multiple-choice format. The answer is considered correct only when it exactly matches the provided answer options. To compute the model's accuracy rate, we initially use regular expressions to match answers and subsequently perform manual inspection and correction of the assessed results.

\paragraph{\bf{The use of thinking templates}}
% % 在主实验中，我们使用了5种思维模板和原始turbo, prompts of thoughts see in Table \ref{pro:tem}。其中，E-KAR使用6种思维模板(origin, Analogical Thinking (AT), Decomposition Thinking 1 (DT), Decomposition Thinking 2 (DST), Plan Thinking (PT), Step Thinking(ST)), Aqua则使用5种(origin, DT, DST, PT, ST)。
% In the main experiment, we used 5 thinking templates and original turbo, prompts of thoughts see in Table \ref{pro:tem}. Among them, E-KAR uses six thought templates (origin, Analogical Thinking (AT), Decomposition Thinking 1 (DT), Decomposition Thinking 2 (DST), Plan Thinking (PT), Step Thinking (ST)), Aqua uses five thought templates (origin, DT, DST, PT, ST).
% % 无特殊说明的情况，实验中均使用这几种思维模板进行最后的投票。
% Unless otherwise specified, these thinking templates were used in the experiment for final voting.

In the main experiment, we utilized five thinking templates and the original base model (GPT-3.5-turbo). The prompts of each thinking template can be found in Table \ref{pro:tem}. Specifically, \textbf{E-KAR} employed all six thinking templates, including origin, Analogy Thinking (AT), Decomposition Thinking 1 (DT), Decomposition Thinking 2 (DST), Plan Thinking (PT), and Sequential Thinking (ST). For \textbf{AQuA}, we used the other five templates excluding AT.

\paragraph{ \bf Baselines.}
We select conventionally and recently published Augmented LLM baselines for model comparison, which can be briefly categorized into three groups: (1) base LLM model (GPT-3.5-turbo); (2) Augmented LLM based on prompt engineering (Auto-CoT); (3) Augmented LLM based on process optimization (SC).
% 下列无特殊说明中，所有的基线方法都使用了相同的意图增强或者投票方案。
% Unless otherwise specified below, all baseline methods use the same intent augmentation or voting scheme.
To ensure a fair comparison, all baseline methods use the same intention enhancement or voting mechanism.

\begin{itemize}
% [leftmargin=*]
	\item {\textbf{GPT-3.5-turbo}}:  % 它是由openai提供的模型。我们在 GPT-3.5 系列中功能最强大且最具成本效益的模型是 GPT-3.5-turbo，它已针对聊天进行了优化，但也适用于传统的完成任务。
    % GPT-3.5-turbo is the most capable and cost-effective model in the GPT-3.5 family provided by openai, which has been optimized for chat but works well for traditional completions tasks as well.
        A base LLM model that builds upon the GPT-3 architecture and incorporates additional training data and techniques to improve performance on natural language processing tasks. It achieves state-of-the-art results on several benchmark datasets, demonstrating its effectiveness in language modeling and downstream tasks. The temperature is set to 0 in the experiment.
        % 实验中temperature设置为0.
        % \item {\textbf{COT}} 
        % \item {\textbf{React}} React \cite{yao2022react} discusses the use of LLMs (large language models) to generate both reasoning traces and task-specific actions in an interleaved manner. This approach allows for better coordination between the two, as reasoning traces help the model to plan and update actions, while actions enable it to interact with external sources and gather more information.
        \item {\textbf{Auto-CoT} \cite{zhang2022automatic}} 
        % Auto-CoT \cite{zhang2022automatic} describes a proposed method called Auto-CoT, which is an automatic CoT (Chain of Thought) prompting method. This method involves sampling questions with diversity and generating reasoning chains to construct demonstrations.
        An automatic CoT (Chain of Thought) prompting method. This approach involves sampling diverse questions and generating reasoning chains to construct demonstrations.
        \item {\textbf{SC} \cite{wang2022self}} 
        % \cite{wang2022self} proposes a new decoding strategy called self-consistency for chain-of-thought prompting. It samples diverse reasoning paths and selects the most consistent answer by marginalizing out the sampled paths. This approach acknowledges that complex reasoning problems have multiple correct ways of thought.
        A new decoding strategy called self-consistency for chain-of-thought prompting. It samples diverse reasoning paths and selects the most consistent answer by marginalizing the sampled paths. This approach acknowledges that complex reasoning problems have multiple correct ways of thought.
        % \item {\textbf{OlaGPT}} 
\end{itemize}

{\renewcommand{\arraystretch}{1.2}
\begin{table*}[ht]
   
	\caption{Performance comparison of different methods on two datasets. }
	\label{tab:all_result}
	\centering
% 	\footnotesize
	\begin{tabular}{@{}c|c|c|c|c|c|c|c|c|c@{}}
		\toprule
		%\hline
		\multirow{2}{*}{Type} & Datasets &\multicolumn{4}{c|}{AQuA}&\multicolumn{4}{c}{E-KAR (Chinese)} \\
		\cline{2-10} & $notes\_retrieval\_type$  & zero-shot & $random$ & $retrieval$ & $combine$ & zero-shot & $random$ & $retrieval$ & $combine$ \\
		\midrule 
            \multirow{3}{*}{baselines} & turbo & 0.3228 & 0.5236 & 0.5315 & 0.5039 &  0.3762 & 0.3731 & 0.403 & 0.3701\\
             & auto\_cot & - &  0.5748 & - & - & - & 0.3791 & - & - \\ 
            & sc & 0.3189 & 0.6142 & 0.5394 & 0.5906 & 0.3761 & 0.4179 & 0.4119 & 0.3851 \\ 
            % & sc & 0.3228 & 0.5945 & 0.5354 & 0.5827 & 0.3731 & 0.4090 & 0.3552 & 0.3881 \\ 
            \midrule
            \multirow{5}{*}{thinking templates} & AT & - & - & - & - & 0.3851 & 0.3821 & 0.3910 & 0.3881 \\
            & DT & 0.5591 & 0.5866 & 0.5827 & 0.5945 & 0.3612 & 0.4119 & 0.4269 & 0.3881 \\
            & DST & 0.5079 & 0.5236 & 0.5984 & 0.5945 & 0.3552 & 0.4030 & 0.3761 & 0.4000 \\
            & PT & 0.5512 & 0.5472 & 0.5827 & 0.5512 & 0.3851 & 0.3552 & 0.4119 & 0.4060 \\
            & ST & 0.5197 & 0.5787 & 0.6102 & 0.5669 & 0.3373 & 0.4179 & 0.4119 & 0.4388 \\
            % & \%Improv.&  \\
            \midrule

           \multirow{2}{*}{Our Framework} & OlaGPT-regex-vote & 0.5945 & \textbf{0.6496} & \textbf{0.6732} & 0.6772 & \textbf{0.4209} & \textbf{0.4597} & \textbf{0.4567} & \textbf{0.4716} \\
           & OlaGPT-llm-vote & \textbf{0.5984} & 0.6417 & 0.6654 & \textbf{0.7047} & 0.4000 & 0.4418 & 0.4358 & 0.4507 \\
           & \%Improv.& 85.38\% & 5.76\% & 24.81\%  & 19.32\% & 11.82\% & 10.00\% & 10.88\% & 22.46\%  \\
            
		\midrule
		\bottomrule
	\end{tabular}
\end{table*}
}

% \subsection{The Model Performance Comparison with Zero-shot Prompts}

% [[[[[[[主实验]]]]]]]
\subsection{The Model Performance Comparison}
% 为了验证我们设计的增强大模型系统对推理任务的有效性，我们对两大类推理数据集上进行了全面的实验比较。表\ref{tab:all_result} 总结了实验结果，其中最佳性能以粗体显示，Improv 表示相对于最佳基线的改进。从实验结果来看，我们有几个发现：
% In order to verify the effectiveness of our designed enhanced LLMs framework for inference tasks, we conduct comprehensive experimental comparisons on two types of inference datasets. Table \ref{tab:all_result} summarizes the experimental results, where the best performance is shown in bold and Improv indicates the improvement over the best baseline. From the experimental results, we have several findings:
In order to evaluate the effectiveness of our augmented LLM framework for inference tasks, we conducted comprehensive experimental comparisons on two types of inference datasets. The results of the experiments are summarized in Table \ref{tab:all_result}. The best performance is indicated in bold, and "Improv" indicates the improvement over the best baseline. Our experiments yielded several findings:

\begin{itemize}
% [leftmargin=*]
   \item {} 
   % The effect of sc is better than turbo, showing that improving certain ensembles is effective.
   The performance of SC is better than GPT-3.5-turbo, suggesting that employing ensemble methods to a certain extent can indeed contribute to enhancing the effectiveness of large-scale models.
   % sc的效果比turbo要好，说明一定的ensembles方式的提升是有效的。
   \item {} 
   % The performance of our method outperforms the results of SC (SC adopts the same regular majority voting as our method), which can explain the effectiveness of our thinking template strategy to a certain extent. The results of different thinking templates are quite different, and voting at different thought templates is better than running multiple rounds and then voting directly.
   The performance of our approach surpasses that of SC (SC adopts the same majority voting extracted by employing regular expressions as our method), which, to a certain extent, demonstrates the effectiveness of our thinking template strategy. The answers of different thinking templates exhibit considerable variation, and conducting voting at different thinking templates ultimately yields better results than simply running multiple rounds and voting.
   % 我们方法的性能优于SC的结果（SC跟我们采取同样的正则多数表决），一定程度上可以说明我们的思维模板策略的有效性。不同思维模板的结果差异较大，最终再进行投票会比直接跑多轮再投票效果要好。
   \item {} 
   % 对比思维模版而言，不同思维模版的效果有所差异。相对而言，ST，DT效果较优。可能是因为这类按步骤解答的方案可能更适合推理类型题。
   % The effects of different thinking templates are different as shown in Table \ref{tab:all_result}. Relatively speaking, ST and DT have better effects. It is because this kind of step-by-step solution may be more suitable for reasoning-type questions.
   The effects of different thinking templates are different, as shown in Table \ref{tab:all_result}. Relatively speaking, ST and DT have better effects. This can be attributed to the fact that this kind of step-by-step solution may be more suitable for reasoning-type questions.
   \item {} 
   % 有主动学习模块（即表中random, retrieval, combine列的结果）加持的结果显著好于zero shot的结果。说明利用hard case作为笔记库的方法可行。
   % The results supported by Active Learning module (That is, the results of the random, retrieval, and combine columns in the Table \ref{tab:all_result}) are significantly better than the results of zero shot. It shows that the method of using hard cases as a note library is feasible.
   The results presented in Table \ref{tab:all_result} demonstrate that the Active Learning module yields significantly better performance compared to the zero-shot approach. Specifically, the random, retrieval, and combine columns exhibit superior performance. These findings suggest that incorporating challenging cases as a note library is a viable strategy.
   \item {} 
   Different Retrieval schemes work differently on different datasets. Overall, the Combine strategy works better.
   % 不同的检索方案在不同数据集上效果不同。总体而言，combine策略效果较优秀。
   \item {} Our method is significantly better than other solutions, thanks to the rational design of the overall framework, the specific reasons are as follows: 1) The effective design of the Active Learning module; 2) The thinking template has achieved the adaptation of different models as expected, and the results under different thinking templates are different; 3) The Controller module plays a good control role and selects the content that better matches the required content; 4) The way of ensembles with different thinking templates designed by the Voting module is effective.
   % 我们的方法显著好于其他方案，得益于整体框架的合理设计，具体原因如下：1）主动学习模块的有效设计；2）思维模版如预期达到了适配不同模型，不同思维模板下结果差异较大；3）控制器模块起到良好控制作用，选择了与所需内容更为匹配的内容，4）Voting模块设计的不同思维模板进行ensembles的方式有效。
\end{itemize}

% [[[[[[[消融实验]]]]]]]
\subsection{Ablation Study}
% 为了探究每个子模块的效果，我们对模型进行消融分析。
% To explore the effect of each sub-module, we perform some ablation analysis on the model. 
To investigate the impact of each sub-module, we conducted an ablation analysis on our framework.

\begin{figure*}[ht]
  \centering
  \includegraphics[width=1\linewidth]{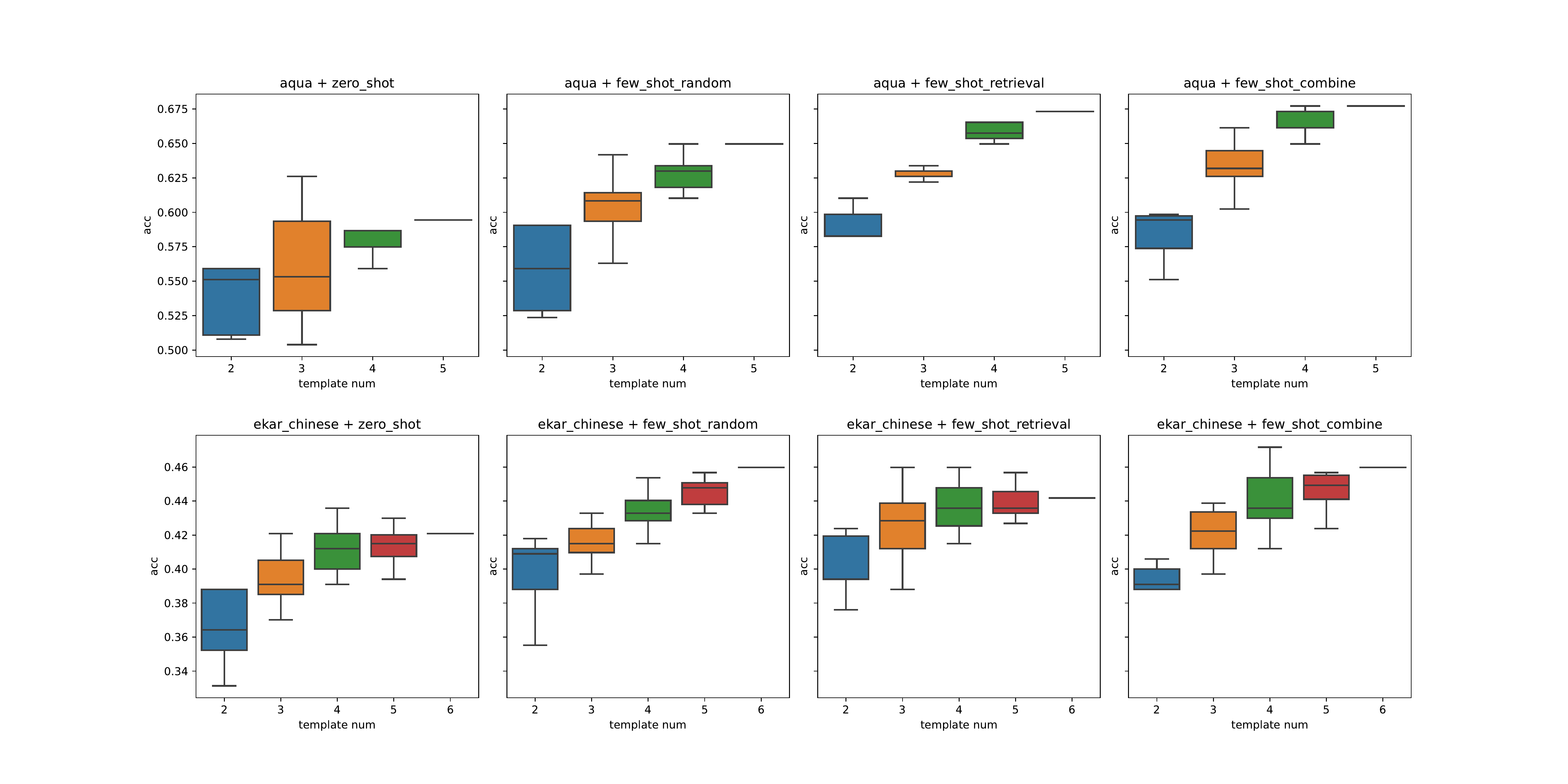}
  \caption{ The performance under the different number of templates and notes retrieval modes.}
  % \Description{A figure shows the overall structure of the OlaGPT model.}
  \label{fig:few_shot}
\end{figure*}

% [[[[[[[主动学习模块消融]]]]]]]
\paragraph{ \bf the Performance of Active Learning module.}
% 为验证主动学习模块的有效性，我们采用是否采取这个模块的方式进行对比。可以看到，few shot 的结果均好于zero shot。直观来看，这种将hard case的形式召回的方式也比直接选择的题效果要好。【原因解释】将大模型相似薄弱的题目作为提示，可以显著提升性能。
In order to verify the effectiveness of the Active Learning module, we compare whether to use this module or not.
% It can be seen that the results of the few-shot are better than zero-shot. 
% Intuitively, this method of recalling the hard case form is also better than the direct selection of questions.
The results shown in Fig \ref{fig:few_shot} indicate that the notes' performance surpasses that of the zero-shot.
% \red{}
% Intuitively, the method of recalling hard cases proves more effective than directly selecting questions.
Taking the similarly weak topics of large models as hints can significantly improve performance.
% 原因易于理解，依据人类的学习过程是对错误经验的积累，我们将相似题目的正确解题过程给到大模型，让它理解后模仿解题是可以让模型学会。
% The reason is easy to understand. According to the human learning process is the accumulation of wrong experiences, we transfer the correct problem-solving process of similar problems to LLMs to let it understand and then imitate the problem-solving.
The underlying rationale is straightforward: humans learn through accumulating experiences, especially incorrect ones. By providing LLM with the correct problem-solving process for similar problems and allowing it to understand and imitate the process, we can enable the LLM to learn from past mistakes.

\begin{figure*}[ht]
  \centering
  \includegraphics[width=1\linewidth]{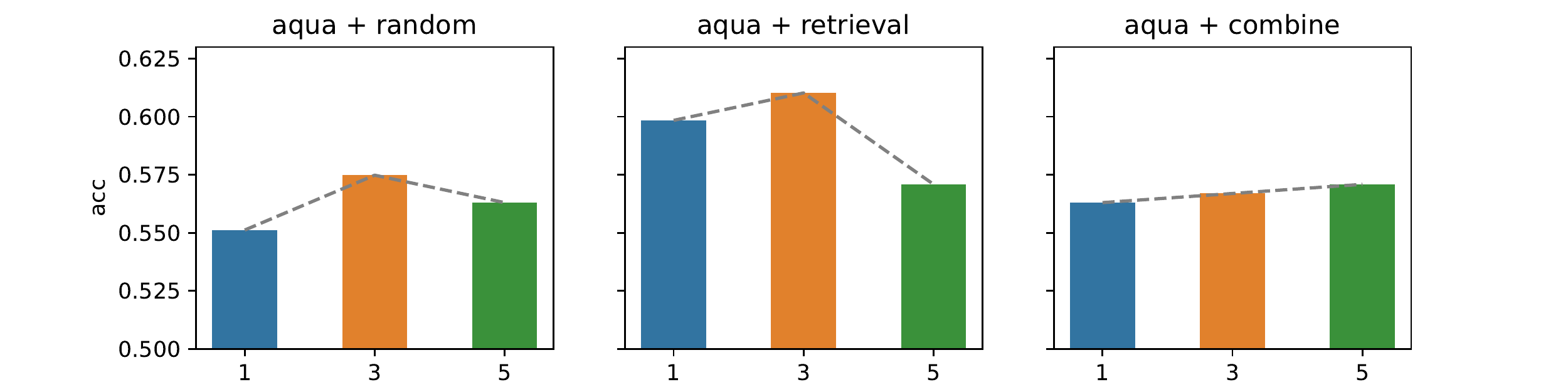}
  \caption{ The performance of ST under the different number of different notes retrieval examples.}
  % \Description{A figure shows the overall structure of the OlaGPT model.}
  \label{fig:few_shot_num}
\end{figure*}

% [[[[[[[关于不同检索方式的讨论]]]]]]]
\paragraph{ \bf the Performance of different retrieval strategies in Controller module.}
% 针对不同的笔记检索策略，我们也进行了实验探究，实验结果如图所示。
For different notes retrieval strategies, we also conducted experiments and the experimental results are shown in Fig \ref{fig:few_shot} and Fig \ref{fig:few_shot_num}. 
% 不同的错题检索策略
% Aqua： combine>retrieval>random>zeroshot
% 观察表\ref{}中的不同$few_shot$值的内容，可以看到在aqua中，$few_shot=3$效果最优，$few_shot=2$次之, 然后是$few_shot=1$, 最差的是zeroshot。

From the results of different $few\_shot\_type$ values in the table \ref{fig:few_shot}, we can see that: the $combine$ strategy has the best effect, Random is the second one, and the worst one is zero-shot.
% 前文介绍到，retrieval和combine两种策略在检索题目类型之后都进行了第二次的检索。retrieval第二次中采用相似度检索而combine则采用随机检索。combine好于retrieval是因为随机检索引入了更大的多样性。对比随机检索来说，相似度检索其实无法一定满足题目类型更加相似，只能是字符上的相似。这一点结论与Auto-CoT的结论相似。最本质的还是需要尽可能找到与提问问题最相关的问题。
Intuitively, finding the most similar question type as an example reference has the greatest gain for LLMs. Due to the large scale of the notes and the large number of questions depending on the type of questions, it is unrealistic to introduce them all as few shots. Both $retrieval$ and $combine$ strategies perform a second search after retrieving the question type. The $retrieval$ strategy uses similarity search the second time and retrieval uses random search. The $combine$ strategy is better than $retrieval$ because random retrieval introduces greater diversity. Compared with random retrieval, similarity retrieval cannot necessarily satisfy the similarity of topic types, only the similarity in characters. This conclusion is similar to that of Auto-CoT. The most essential thing is to find the most relevant questions.

The best result of \textbf{E-KAR} (chinese) is the $random$ strategy. There are several reasons here: 1) The types of questions are not detailed enough. Because of this kind of analogical reasoning questions, the type of questions generated by the model is almost an analogy question, unlike the question types of math questions that can be more detailed. 2) Templates are not customizable. Except for AT for analogy thinking, other thinking templates are made for mathematics questions.
% ekar（chinese）的数据集则random最优。这里有几个原因：1）题目类型不够细致。因这种类比推理题，模型生成的题目类型几乎为类比问题，不像数学题的题目类型能做到较为细致。2）模板非定制化。我们模板除AT之外是为类比思想定制化之外，其他的思维模板为数学类题目所制作，并非解决类比推理的较好思路。3）

% [[[[[[[探讨不同思维模板的性能]]]]]]]
\paragraph{ \bf the Performance of Different Thoughts in Reasoning module.}
% 思想模板的效果：在此小节主要探讨下不同思维模版在不同任务的效果。
This section mainly discusses the effect of different thinking templates on different tasks.
% 从表\ref{tab:all_result}中可以发现不同的思维模板在不同数据集和不同检索策略下的效果略有差异，但都比turbo结果要好。这一点也验证了文中对于一个模板无法良好解决多种任务的认知。目前aqua上效果较好的前三有ST-retrieval, DST-retrieval, DT-combine。E-KAR上效果较好的则有ST-combine, DT-retrieval, DST-random。
% From the table \ref{tab:all_result}, it can be found that different thinking templates have slightly different effects under different data sets and different retrieval strategies, but they are all better than turbo results, which also verifies the cognition in the article that one template cannot solve multiple tasks well. At present, the top three with better effects on aqua are ST-retrieval, DST-retrieval, and DT-combine. The better effects on E-KAR include ST-combine, DT-retrieval, and DST-random. 
From Table \ref{tab:all_result}, it is evident that different thinking templates exhibit varying effects across distinct datasets and retrieval strategies. However, all of them surpass the turbo's results, thereby validating the notion presented in the article that a single template cannot effectively address multiple tasks. Currently, the top three strategies yielding superior results on \textbf{AQuA} are ST-retrieval, DST-retrieval, and DT-combine. For \textbf{E-KAR}, the more effective strategies include ST-combine, DT-retrieval, and DST-random.

% 经过后验分析，所示部分模板最终的正确率差异不大（如表\ref{tab:all_result}所示，aqua不同模板的最优结果几乎在0.58-0.61），但他们的一致性并不高（如表\ref{tab:consistency}所示，）。正是这个原因，通过设计vote模块能尽可能选对更多的正确答案。
% 【】
% \red{}
After posterior analysis, the final correct rate of some templates has little difference as shown in the table \ref{tab:all_result}, but their consistency is not high. For this very reason, designing a voting module can help select as many correct answers as possible.
% 在附录中我们还分析了不同vote策略下的准确率峰值。
% We also analyze the peak accuracy under different voting strategies in Table \ref{tab:high} In the appendix.
We have also examined the possible maximum accuracy that can be achieved by different voting strategies. Please refer to Table \ref{tab:high} in the appendix.

% 
% 执行：单模板的在vote之前的效果
% 【放入附录中，回答有效数】将number=[1-6]的比例全加起来部分没到百分百的原因是部分解析错误所致。

{\renewcommand{\arraystretch}{1.2}
\begin{table}[ht]
   
	\caption{Consistency analysis of the thinking templates.}
	\label{tab:consistency}
	\centering
% 	\footnotesize
\resizebox{\linewidth}{!}{ 
	\begin{tabular}{@{}c|c|c|c|c|c|c|c|c@{}}
		\toprule
		%\hline
		\multirow{2}{*}{\#consistency} &\multicolumn{4}{c|}{AQuA}&\multicolumn{4}{c}{E-KAR (Chinese)}   \\
		\cline{2-9} & zero-shot & $random$ & $retrieval$ & $combine$ & zero-shot & $random$ & $retrieval$ & $combine$ \\
		\midrule 
            c=2 & 66 & 56 & 45 & 61 & 42 & 45 & 44 & 40 \\
            c=3 & 95 & 71 & 79 & 73 & 94 & 94 & 91 & 92 \\
            c=4 & 63 & 48 & 63 & 61 & 102 & 82 & 81 & 86 \\
            c=5 & 29 & 78 & 67 & 57 & 64 & 74 & 71 & 75 \\
            c=6 & - & - & - & - & 16 & 16 & 16 & 16 \\
		\midrule
		\bottomrule
	\end{tabular}
 }
\end{table}
}

% [[[[[[[探讨投票模块的性能]]]]]]]
\paragraph{ \bf the Performance of Voting module.} We explored different voting strategies and the experimental situation of voting under different template numbers. In this experiment, if it is not specified, the regex results will be used for discussion.

\begin{figure*}[ht]
  \centering
  \includegraphics[width=1\linewidth]{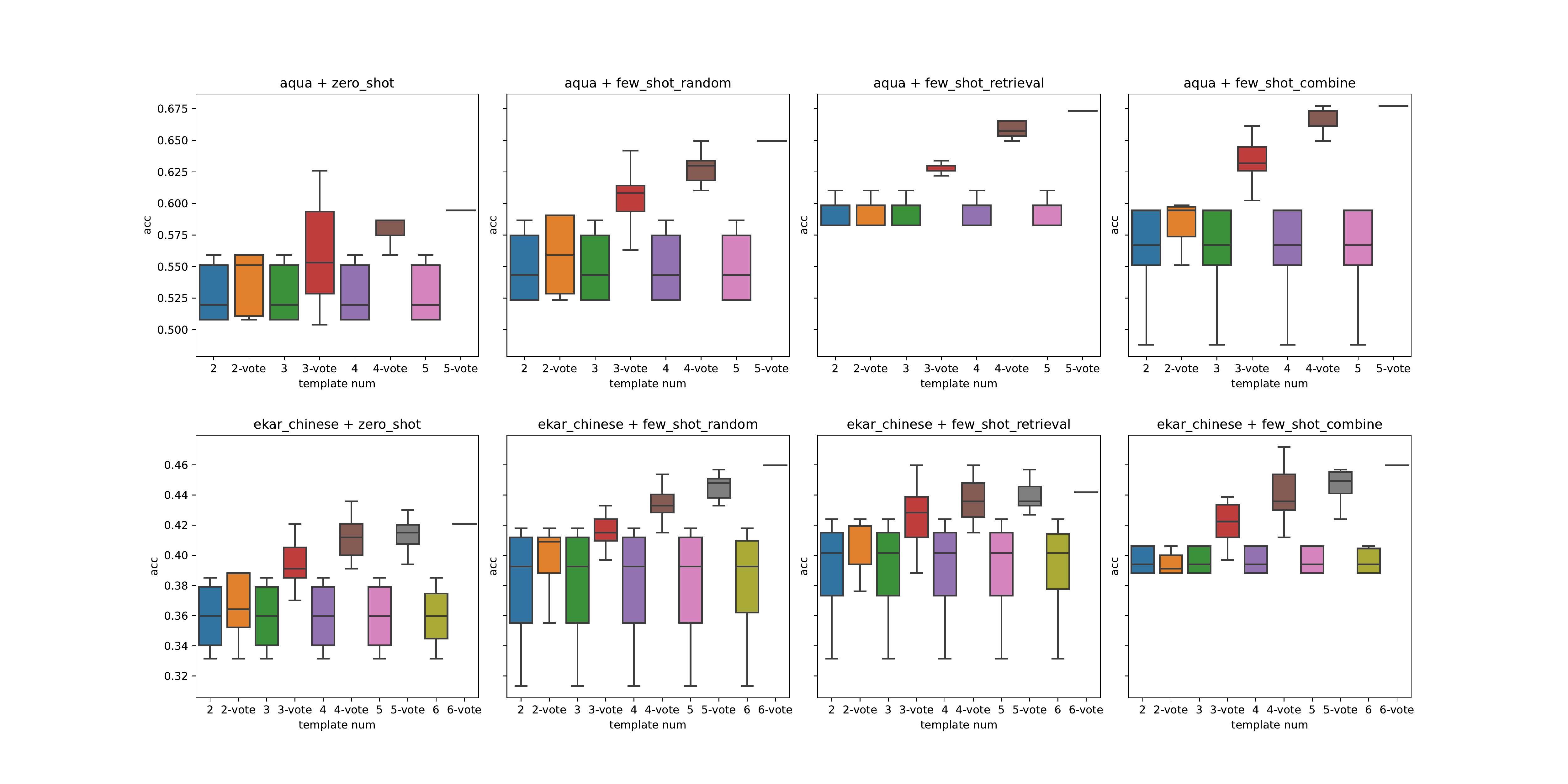}
  \caption{ The performance before and after voting.}
  % \Description{A figure shows the overall structure of the OlaGPT model.}
  \label{fig:vote_box}
\end{figure*}

\begin{itemize}
% [leftmargin=*]
   \item {} % 【对比vote前的效果】如图\ref{fig:vote_box}所示，对比每个单模板的效果几乎都有所增强。可以说明vote这种提升方式是切实可行的。
    % As shown in Figure \ref{fig:vote_box}, the effect of comparing every single template is almost enhanced. It can be shown that the promotion method of vote is feasible.
    As shown in Fig \ref{fig:vote_box}, the performance of the model after voting is significantly improved compared to that of individual thinking templates, which also validates the effectiveness of the voting approach.
    % 【原因】不同模板适合不同题型，vote策略能集百家所长，进一步提升框架性能。我们将平票算为全部正确的结果作为上界，算为全部错误的结果算为下界，具体数值如附录中的]表所示ref{tab:high}。【待完善】
    % Different templates are suitable for different question types, and the vote strategy can integrate the strengths of hundreds of schools to further improve the performance of the framework. We take all correct results as the upper bound, and all wrong results as the lower bound. The specific values are shown in the Table \ref{tab:high} in the Appendix.
    Different thinking templates are appropriate for different types of questions, and the voting strategy can consolidate the strengths of different thinking templates to further enhance the framework's performance. We consider all correct results as the upper bound and all incorrect results as the lower bound with the regex-vote (majority vote by regex expression) method. The specific values are presented in Table \ref{tab:high} in the Appendix.
    \item {} % 【不同模板数下】如图\ref{fig:vote_box}，随着模版的数量增加，准确率增加。
    % As shown in \ref{fig:vote_box}, as the number of templates increases, the accuracy rate increases.
    As shown in Fig \ref{fig:few_shot}, the accuracy rate increases with the growth in the number of templates.
    % % 这一点可以看附录中正则-upper最多模板数下与三模板的比较，多模板数下的正则天花板更高。
    % \red{This point can be seen in the appendix in the comparison between regex-upper with the largest number of templates and three templates. The ceiling of regularization with multiple templates is higher.}
    % % 【原因】随着模板数的增加，模板之间正确的个数交叠情况或许会随之增加，选项也可能更加丰富，由此给最终的vote模块带来更大的正确率空间。
    % As the number of templates increases, the number of correct overlaps between templates may increase, and the options may also be more abundant, thus bringing greater accuracy to the final vote module.
    As the number of templates increases, the variety of answer combinations generated by different templates also expands, thereby offering the voting module a broader potential upper limit for performance improvement.

    \item {} % 【对比大模型和正则的vote】正则的效果能保证多数投票机制，而大模型vote可以选出非票数最多的为正确的情况。
    % The effect of regex-vote can guarantee the majority voting mechanism, and the llm-vote can select the correct case with the largest number of non-votes.
    The regex-vote strategy can ensure the majority voting mechanism, while the llm-vote strategy relies on the explanation of the candidate's answers, which may have a larger degree of uncertainty.
    % 【原因】例如在5模板结果的情况下，大模型能有概率把3错2对的情况给选对，而这是正则最多化Vote做不到的。
    % For example, in the case of all template results, the llm-vote can have the probability to select the correct case of 3 wrong and 2 right, but this is not possible with the majority regex-vote.
    Consider an example with five templates, where there are 2 correct and 3 incorrect answers. The regex-vote strategy will always select the incorrect. The llm-vote strategy still has the possibility of selecting the correct answer, based on the different confidence of weights on different answers. Consequently, the potential accuracy upper bound of the llm-vote strategy may be even higher.
    
\end{itemize}

\begin{figure*}[ht]
  \centering
  \includegraphics[width=1\linewidth]{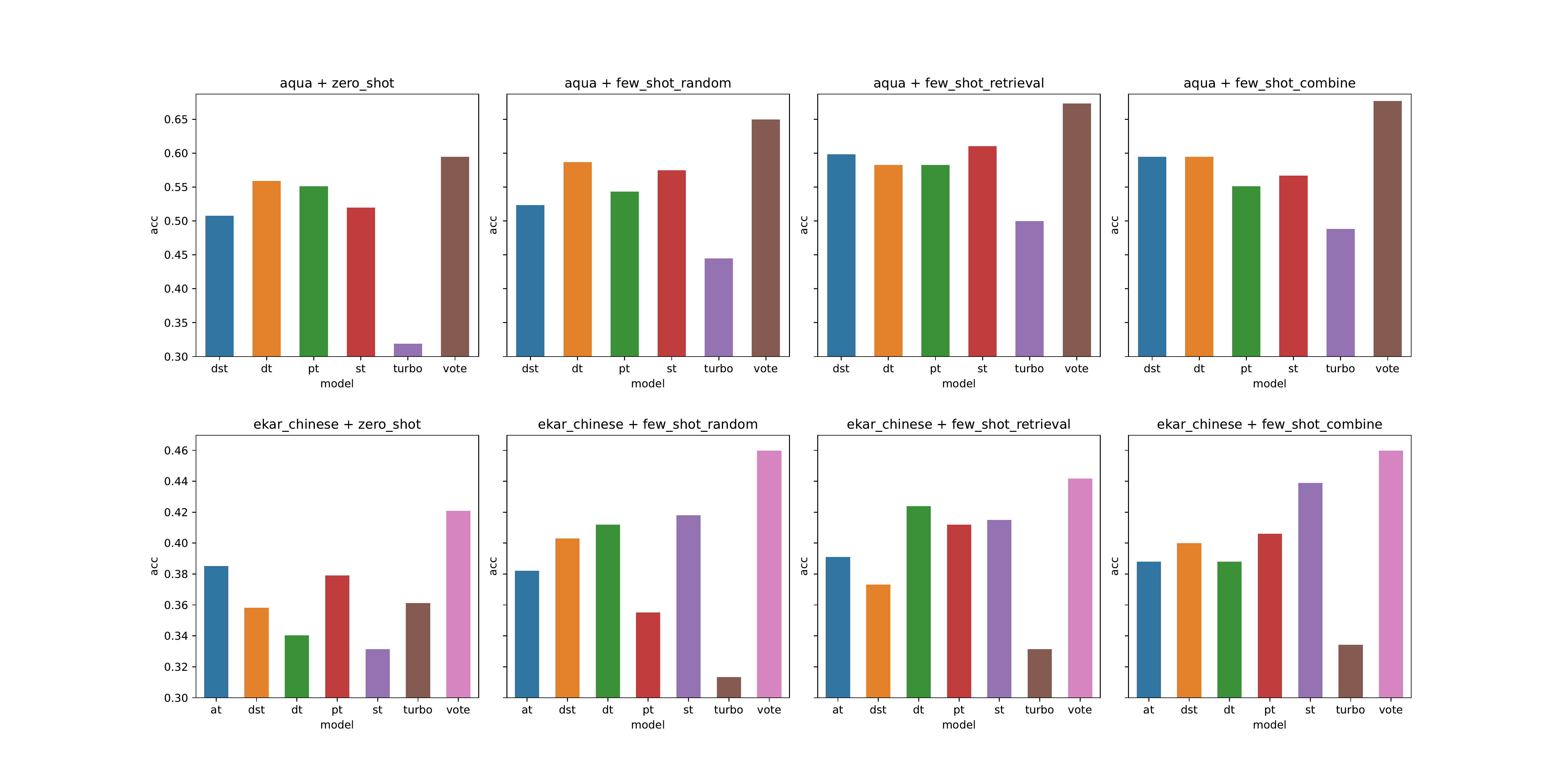}
  \caption{ The performance before and after voting.}
  % \Description{A figure shows the overall structure of the OlaGPT model.}
  \label{fig:vote_bar}
\end{figure*}

% 综上所述，我们在以前的模型之上增量添加一个新的子模块或特征时，我们都可以观察到整体推荐性能的提高，这说明了主动学习、控制、思维和投票模块的有效性。
In summary, we can observe the improvement of the overall recommendation performance whenever we incrementally add a new module or feature on top of the previous model, which illustrates the effectiveness of the Active Learning, Controller, Thought, and Voting modules.

\subsection{Hyperparameter Study}
% 在此小节中我们对模型中的超参数进行了一些研究。
% 错题阈值的参数实验：为了探究notes作为few-shot的个数多少更合适，我们设置了例题的个数实验。以aqua数据集在较好的ST结果模板下的实验结果为例，数据结果如图\ref{fig:few_shot_num}所示。可以发现不同检索策略下的个数最优值不同。combine策略稳步提升，而random和retrieval则先升后降。更多样例个数的时候可能引入更多噪声，寻找少量最合适的相似类型题即可。
% In this subsection we do some research on the hyperparameters in the model. In order to explore how many notes are more appropriate as the number of few-shots, we set up an experiment on the number of sample questions. Take the experimental results of the aqua dataset under the better ST result template as an example, the results are shown in Figure \ref{fig:few_shot_num}. It can be found that the optimal value of the number under different retrieval strategies is different. The combine strategy improves steadily, while the random and retrieval first increase and then decrease. When there are more samples, more noise may be introduced, just find a small number of the most suitable similar types of questions.
% [1, 3, 5]

In this subsection, we investigate the model's hyperparameters. To determine the optimal number of notes to use as the example reference, we conduct an experiment on the number of examples. Taking the experimental results of the \textbf{AQuA} dataset under the ST template as an example, the outcomes are displayed in Fig \ref{fig:few_shot_num}. It can be found that the optimal value for the number of notes varies among different retrieval strategies. The $combine$ strategy exhibits consistent improvement, while the $random$ and $retrieval$ strategies first increase and then decrease. When more sample questions are included, additional noise may be introduced. Thus, it is important to find the appropriate number of examples, neither too few nor too many, making a trade-off.

\section{Conclusion}
% 本文主要设计完成一个基于人类认知的增强LLM框架（OlaGPT），重点解决较难的推理问题。具体地，参考人类认知理论，OlaGPT包含了六大模块：意图增强、记忆、主动学习、思维、控制者和投票模块。
This paper designs an enhanced LLM cognition framework (OlaGPT), aiming to solve difficult reasoning problems with human-like problem-solving abilities. Specifically, referring to the theory of human cognition, OlaGPT proposes to approximate cognitive modules, such as attention (for entention enhancement), Memory, Learning, Reasoning, action selection (Controller) and decision making.
% 用户的提问经过意图增强模块后获得更精细的表达，控制者模块控制利用此表达选择所需的仓库内容，完成多个思维模板的填充。获取异步执行多个思维模板的结果之后利用vote模块获得更鲁棒的效果。
% After the user's question is passed through the Intention Enhancement module, a more refined prompt is obtained.
The user's query undergoes refinement through the Intention Enhancement module, achieving a more precise expression. 
% The Controller module utilizes this expression to select the required content from the repository, completing the instantiation of multiple reasoning templates. 
Then, the Controller module controls and utilizes this expression to select the required library content, and completes the filling of multiple thinking templates.
% Use the Voting module to obtain more robust results after obtaining the results of executing multiple thinking templates asynchronously.
After acquiring the results of asynchronously executing multiple reasoning templates, a more robust effect is achieved using the Voting module.
% we employs the voting module to achieve more robust effects.
% 我们在两个真实推理数据集上进行了实验，结果表明OlaGPT方法在推理问题的解答上优于现有方法的性能。此外，我们还证明了模型中每个部分设计的有效性。模块大部分设计可插拔，不同场景可依需求来确定所需模块。
We conduct experiments on two real inference datasets and show that the OlaGPT method outperforms existing methods in answering inference questions. In addition, we also demonstrate the effectiveness of the design of each part in the model. Most of the modules are designed to be pluggable, and the required modules can be determined according to the needs of different scenarios.

% 【未来工作】
% 在后续工作中，我们会持续优化完善每个子模块的功能，预计完成一个具有人类思维能力的易用增强大模型框架。
In the follow-up work, we will continue to optimize and improve the functions of each sub-module, and it is expected to complete an easy-to-use enhanced large-scale model framework with human thinking ability.
% 首先会增加更多样化的数据集和baselines进行实验测试。此外，我们将持续优化每个子模块的设计，进行更多实验作证设想。
First, more diverse datasets and baselines will be added for experimental testing. In addition, we will continue to optimize the design of each sub-module and conduct more experiments to testify our ideas.
% 不同模型和数据类型可能会
% 后续工作
% \textcolor{green}{
% % 此外，我们目前在尝试打造一个自动化的意图增强模块，设置种子数据集和相关的instruct，调用gpt接口生成一批训练数据。利用开源的llama模型和lora技术，使用指令生成的数据进行微调。意图实现一个自动化用户输入增强的模块。此模块还在开发实验中，后续如有效果会放出相关的实验结果。
% Additionally, we are currently trying to build an automated intent enhancement module, set up a seed dataset and related instructions, and call the gpt interface to generate a batch of training data. Using the open source llama model and lora technology, fine-tuning is performed using the data generated by the instruction. Intended to implement a module that automates user input enhancements. This module is still under development and experimentation, and relevant experiment results will be released in the future if there is an effect.}

\begin{acks}
\end{acks}

\bibliographystyle{ACM-Reference-Format}
\bibliography{main_bib}


\begin{thebibliography}{36}

%%% the Web address.

\ifx \showCODEN    \undefined \def \showCODEN     #1{\unskip}     \fi
\ifx \showDOI      \undefined \def \showDOI       #1{#1}\fi
\ifx \showISBNx    \undefined \def \showISBNx     #1{\unskip}     \fi
\ifx \showISBNxiii \undefined \def \showISBNxiii  #1{\unskip}     \fi
\ifx \showISSN     \undefined \def \showISSN      #1{\unskip}     \fi
\ifx \showLCCN     \undefined \def \showLCCN      #1{\unskip}     \fi
\ifx \shownote     \undefined \def \shownote      #1{#1}          \fi
\ifx \showarticletitle \undefined \def \showarticletitle #1{#1}   \fi
\ifx \showURL      \undefined \def \showURL       {\relax}        \fi
% invisible to TeX
\providecommand\bibfield[2]{#2}
\providecommand\bibinfo[2]{#2}
\providecommand\natexlab[1]{#1}
\providecommand\showeprint[2][]{arXiv:#2}

\bibitem[Anderson(2009)]%
        {anderson2009can}
\bibfield{author}{\bibinfo{person}{John~R Anderson}.}
  \bibinfo{year}{2009}\natexlab{}.
\newblock \bibinfo{booktitle}{\emph{How can the human mind occur in the
  physical universe?}}
\newblock \bibinfo{publisher}{Oxford University Press}.
\newblock


\bibitem[Anderson and Lebiere(2014)]%
        {anderson2014atomic}
\bibfield{author}{\bibinfo{person}{John~R Anderson} {and}
  \bibinfo{person}{Christian~J Lebiere}.} \bibinfo{year}{2014}\natexlab{}.
\newblock \bibinfo{booktitle}{\emph{The atomic components of thought}}.
\newblock \bibinfo{publisher}{Psychology Press}.
\newblock


\bibitem[Brown et~al\mbox{.}(2020)]%
        {brown2020language}
\bibfield{author}{\bibinfo{person}{Tom Brown}, \bibinfo{person}{Benjamin Mann},
  \bibinfo{person}{Nick Ryder}, \bibinfo{person}{Melanie Subbiah},
  \bibinfo{person}{Jared~D Kaplan}, \bibinfo{person}{Prafulla Dhariwal},
  \bibinfo{person}{Arvind Neelakantan}, \bibinfo{person}{Pranav Shyam},
  \bibinfo{person}{Girish Sastry}, \bibinfo{person}{Amanda Askell},
  {et~al\mbox{.}}} \bibinfo{year}{2020}\natexlab{}.
\newblock \showarticletitle{Language models are few-shot learners}.
\newblock \bibinfo{journal}{\emph{Advances in neural information processing
  systems}}  \bibinfo{volume}{33} (\bibinfo{year}{2020}),
  \bibinfo{pages}{1877--1901}.
\newblock


\bibitem[Chen et~al\mbox{.}(2022)]%
        {chen-etal-2022-e}
\bibfield{author}{\bibinfo{person}{Jiangjie Chen}, \bibinfo{person}{Rui Xu},
  \bibinfo{person}{Ziquan Fu}, \bibinfo{person}{Wei Shi},
  \bibinfo{person}{Zhongqiao Li}, \bibinfo{person}{Xinbo Zhang},
  \bibinfo{person}{Changzhi Sun}, \bibinfo{person}{Lei Li},
  \bibinfo{person}{Yanghua Xiao}, {and} \bibinfo{person}{Hao Zhou}.}
  \bibinfo{year}{2022}\natexlab{}.
\newblock \showarticletitle{{E}-{KAR}: A Benchmark for Rationalizing Natural
  Language Analogical Reasoning}. In \bibinfo{booktitle}{\emph{Findings of the
  Association for Computational Linguistics: ACL 2022}}.
  \bibinfo{publisher}{Association for Computational Linguistics},
  \bibinfo{address}{Dublin, Ireland}, \bibinfo{pages}{3941--3955}.
\newblock
\urldef\tempurl%
\url{https://aclanthology.org/2022.findings-acl.311}
\showURL{%
\tempurl}


\bibitem[Chen et~al\mbox{.}(2021)]%
        {chen2021evaluating}
\bibfield{author}{\bibinfo{person}{Mark Chen}, \bibinfo{person}{Jerry Tworek},
  \bibinfo{person}{Heewoo Jun}, \bibinfo{person}{Qiming Yuan},
  \bibinfo{person}{Henrique~Ponde de Oliveira~Pinto}, \bibinfo{person}{Jared
  Kaplan}, \bibinfo{person}{Harri Edwards}, \bibinfo{person}{Yuri Burda},
  \bibinfo{person}{Nicholas Joseph}, \bibinfo{person}{Greg Brockman},
  \bibinfo{person}{Alex Ray}, \bibinfo{person}{Raul Puri},
  \bibinfo{person}{Gretchen Krueger}, \bibinfo{person}{Michael Petrov},
  \bibinfo{person}{Heidy Khlaaf}, \bibinfo{person}{Girish Sastry},
  \bibinfo{person}{Pamela Mishkin}, \bibinfo{person}{Brooke Chan},
  \bibinfo{person}{Scott Gray}, \bibinfo{person}{Nick Ryder},
  \bibinfo{person}{Mikhail Pavlov}, \bibinfo{person}{Alethea Power},
  \bibinfo{person}{Lukasz Kaiser}, \bibinfo{person}{Mohammad Bavarian},
  \bibinfo{person}{Clemens Winter}, \bibinfo{person}{Philippe Tillet},
  \bibinfo{person}{Felipe~Petroski Such}, \bibinfo{person}{Dave Cummings},
  \bibinfo{person}{Matthias Plappert}, \bibinfo{person}{Fotios Chantzis},
  \bibinfo{person}{Elizabeth Barnes}, \bibinfo{person}{Ariel Herbert-Voss},
  \bibinfo{person}{William~Hebgen Guss}, \bibinfo{person}{Alex Nichol},
  \bibinfo{person}{Alex Paino}, \bibinfo{person}{Nikolas Tezak},
  \bibinfo{person}{Jie Tang}, \bibinfo{person}{Igor Babuschkin},
  \bibinfo{person}{Suchir Balaji}, \bibinfo{person}{Shantanu Jain},
  \bibinfo{person}{William Saunders}, \bibinfo{person}{Christopher Hesse},
  \bibinfo{person}{Andrew~N. Carr}, \bibinfo{person}{Jan Leike},
  \bibinfo{person}{Josh Achiam}, \bibinfo{person}{Vedant Misra},
  \bibinfo{person}{Evan Morikawa}, \bibinfo{person}{Alec Radford},
  \bibinfo{person}{Matthew Knight}, \bibinfo{person}{Miles Brundage},
  \bibinfo{person}{Mira Murati}, \bibinfo{person}{Katie Mayer},
  \bibinfo{person}{Peter Welinder}, \bibinfo{person}{Bob McGrew},
  \bibinfo{person}{Dario Amodei}, \bibinfo{person}{Sam McCandlish},
  \bibinfo{person}{Ilya Sutskever}, {and} \bibinfo{person}{Wojciech Zaremba}.}
  \bibinfo{year}{2021}\natexlab{}.
\newblock \bibinfo{title}{Evaluating Large Language Models Trained on Code}.
\newblock
\newblock
\showeprint[arxiv]{2107.03374}~[cs.LG]


\bibitem[Diao et~al\mbox{.}(2023)]%
        {diao2023active}
\bibfield{author}{\bibinfo{person}{Shizhe Diao}, \bibinfo{person}{Pengcheng
  Wang}, \bibinfo{person}{Yong Lin}, {and} \bibinfo{person}{Tong Zhang}.}
  \bibinfo{year}{2023}\natexlab{}.
\newblock \showarticletitle{Active Prompting with Chain-of-Thought for Large
  Language Models}.
\newblock \bibinfo{journal}{\emph{arXiv preprint arXiv:2302.12246}}
  (\bibinfo{year}{2023}).
\newblock


\bibitem[Drori et~al\mbox{.}(2022)]%
        {Drori_2022}
\bibfield{author}{\bibinfo{person}{Iddo Drori}, \bibinfo{person}{Sarah Zhang},
  \bibinfo{person}{Reece Shuttleworth}, \bibinfo{person}{Leonard Tang},
  \bibinfo{person}{Albert Lu}, \bibinfo{person}{Elizabeth Ke},
  \bibinfo{person}{Kevin Liu}, \bibinfo{person}{Linda Chen},
  \bibinfo{person}{Sunny Tran}, \bibinfo{person}{Newman Cheng},
  \bibinfo{person}{Roman Wang}, \bibinfo{person}{Nikhil Singh},
  \bibinfo{person}{Taylor~L. Patti}, \bibinfo{person}{Jayson Lynch},
  \bibinfo{person}{Avi Shporer}, \bibinfo{person}{Nakul Verma},
  \bibinfo{person}{Eugene Wu}, {and} \bibinfo{person}{Gilbert Strang}.}
  \bibinfo{year}{2022}\natexlab{}.
\newblock \showarticletitle{A neural network solves, explains, and generates
  university math problems by program synthesis and few-shot learning at human
  level}.
\newblock \bibinfo{journal}{\emph{Proceedings of the National Academy of
  Sciences}} \bibinfo{volume}{119}, \bibinfo{number}{32} (\bibinfo{date}{aug}
  \bibinfo{year}{2022}).
\newblock
\urldef\tempurl%
\url{https://doi.org/10.1073/pnas.2123433119}
\showDOI{\tempurl}


\bibitem[Gao et~al\mbox{.}(2023)]%
        {gao2023pal}
\bibfield{author}{\bibinfo{person}{Luyu Gao}, \bibinfo{person}{Aman Madaan},
  \bibinfo{person}{Shuyan Zhou}, \bibinfo{person}{Uri Alon},
  \bibinfo{person}{Pengfei Liu}, \bibinfo{person}{Yiming Yang},
  \bibinfo{person}{Jamie Callan}, {and} \bibinfo{person}{Graham Neubig}.}
  \bibinfo{year}{2023}\natexlab{}.
\newblock \bibinfo{title}{PAL: Program-aided Language Models}.
\newblock
\newblock
\showeprint[arxiv]{2211.10435}~[cs.CL]


\bibitem[Groves et~al\mbox{.}(2008)]%
        {groves2008linking}
\bibfield{author}{\bibinfo{person}{Kevin Groves}, \bibinfo{person}{Charles
  Vance}, {and} \bibinfo{person}{Yongsun Paik}.}
  \bibinfo{year}{2008}\natexlab{}.
\newblock \showarticletitle{Linking linear/nonlinear thinking style balance and
  managerial ethical decision-making}.
\newblock \bibinfo{journal}{\emph{Journal of Business Ethics}}
  \bibinfo{volume}{80} (\bibinfo{year}{2008}), \bibinfo{pages}{305--325}.
\newblock


\bibitem[Hu et~al\mbox{.}(2021)]%
        {hu2021lora}
\bibfield{author}{\bibinfo{person}{Edward~J Hu}, \bibinfo{person}{Yelong Shen},
  \bibinfo{person}{Phillip Wallis}, \bibinfo{person}{Zeyuan Allen-Zhu},
  \bibinfo{person}{Yuanzhi Li}, \bibinfo{person}{Shean Wang},
  \bibinfo{person}{Lu Wang}, {and} \bibinfo{person}{Weizhu Chen}.}
  \bibinfo{year}{2021}\natexlab{}.
\newblock \showarticletitle{Lora: Low-rank adaptation of large language
  models}.
\newblock \bibinfo{journal}{\emph{arXiv preprint arXiv:2106.09685}}
  (\bibinfo{year}{2021}).
\newblock


\bibitem[Johnson et~al\mbox{.}(2019)]%
        {johnson2019billion}
\bibfield{author}{\bibinfo{person}{Jeff Johnson}, \bibinfo{person}{Matthijs
  Douze}, {and} \bibinfo{person}{Herv{\'e} J{\'e}gou}.}
  \bibinfo{year}{2019}\natexlab{}.
\newblock \showarticletitle{Billion-scale similarity search with gpus}.
\newblock \bibinfo{journal}{\emph{IEEE Transactions on Big Data}}
  \bibinfo{volume}{7}, \bibinfo{number}{3} (\bibinfo{year}{2019}),
  \bibinfo{pages}{535--547}.
\newblock


\bibitem[Kallio(2011)]%
        {kallio2011integrative}
\bibfield{author}{\bibinfo{person}{Eeva Kallio}.}
  \bibinfo{year}{2011}\natexlab{}.
\newblock \showarticletitle{Integrative thinking is the key: An evaluation of
  current research into the development of adult thinking}.
\newblock \bibinfo{journal}{\emph{Theory \& Psychology}} \bibinfo{volume}{21},
  \bibinfo{number}{6} (\bibinfo{year}{2011}), \bibinfo{pages}{785--801}.
\newblock


\bibitem[Khot et~al\mbox{.}(2023)]%
        {khot2023decomposed}
\bibfield{author}{\bibinfo{person}{Tushar Khot}, \bibinfo{person}{Harsh
  Trivedi}, \bibinfo{person}{Matthew Finlayson}, \bibinfo{person}{Yao Fu},
  \bibinfo{person}{Kyle Richardson}, \bibinfo{person}{Peter Clark}, {and}
  \bibinfo{person}{Ashish Sabharwal}.} \bibinfo{year}{2023}\natexlab{}.
\newblock \bibinfo{title}{Decomposed Prompting: A Modular Approach for Solving
  Complex Tasks}.
\newblock
\newblock
\showeprint[arxiv]{2210.02406}~[cs.CL]


\bibitem[Kim and Horii(2016)]%
        {kim2016analogical}
\bibfield{author}{\bibinfo{person}{Eunyoung Kim} {and}
  \bibinfo{person}{Hideyuki Horii}.} \bibinfo{year}{2016}\natexlab{}.
\newblock \showarticletitle{Analogical thinking for generation of innovative
  ideas: An exploratory study of influential factors}.
\newblock \bibinfo{journal}{\emph{Interdisciplinary Journal of Information,
  Knowledge, and Management}}  \bibinfo{volume}{11} (\bibinfo{year}{2016}),
  \bibinfo{pages}{201}.
\newblock


\bibitem[Kokinov(1994)]%
        {kokinov1994hybrid}
\bibfield{author}{\bibinfo{person}{Boicho Kokinov}.}
  \bibinfo{year}{1994}\natexlab{}.
\newblock \showarticletitle{A hybrid model of reasoning by analogy}.
\newblock \bibinfo{journal}{\emph{Advances in connectionist and neural
  computation theory}}  \bibinfo{volume}{2} (\bibinfo{year}{1994}),
  \bibinfo{pages}{247--318}.
\newblock


\bibitem[Kotseruba and Tsotsos(2020)]%
        {kotseruba202040}
\bibfield{author}{\bibinfo{person}{Iuliia Kotseruba} {and}
  \bibinfo{person}{John~K Tsotsos}.} \bibinfo{year}{2020}\natexlab{}.
\newblock \showarticletitle{40 years of cognitive architectures: core cognitive
  abilities and practical applications}.
\newblock \bibinfo{journal}{\emph{Artificial Intelligence Review}}
  \bibinfo{volume}{53}, \bibinfo{number}{1} (\bibinfo{year}{2020}),
  \bibinfo{pages}{17--94}.
\newblock


\bibitem[Lai(2011)]%
        {lai2011critical}
\bibfield{author}{\bibinfo{person}{Emily~R Lai}.}
  \bibinfo{year}{2011}\natexlab{}.
\newblock \showarticletitle{Critical thinking: A literature review}.
\newblock \bibinfo{journal}{\emph{Pearson's Research Reports}}
  \bibinfo{volume}{6}, \bibinfo{number}{1} (\bibinfo{year}{2011}),
  \bibinfo{pages}{40--41}.
\newblock


\bibitem[Laird(2019)]%
        {laird2019soar}
\bibfield{author}{\bibinfo{person}{John~E Laird}.}
  \bibinfo{year}{2019}\natexlab{}.
\newblock \bibinfo{booktitle}{\emph{The Soar cognitive architecture}}.
\newblock \bibinfo{publisher}{MIT press}.
\newblock


\bibitem[Laird et~al\mbox{.}(1987)]%
        {laird1987soar}
\bibfield{author}{\bibinfo{person}{John~E Laird}, \bibinfo{person}{Allen
  Newell}, {and} \bibinfo{person}{Paul~S Rosenbloom}.}
  \bibinfo{year}{1987}\natexlab{}.
\newblock \showarticletitle{Soar: An architecture for general intelligence}.
\newblock \bibinfo{journal}{\emph{Artificial intelligence}}
  \bibinfo{volume}{33}, \bibinfo{number}{1} (\bibinfo{year}{1987}),
  \bibinfo{pages}{1--64}.
\newblock


\bibitem[Ling et~al\mbox{.}(2017)]%
        {ling2017program}
\bibfield{author}{\bibinfo{person}{Wang Ling}, \bibinfo{person}{Dani Yogatama},
  \bibinfo{person}{Chris Dyer}, {and} \bibinfo{person}{Phil Blunsom}.}
  \bibinfo{year}{2017}\natexlab{}.
\newblock \showarticletitle{Program induction by rationale generation: Learning
  to solve and explain algebraic word problems}.
\newblock \bibinfo{journal}{\emph{arXiv preprint arXiv:1705.04146}}
  (\bibinfo{year}{2017}).
\newblock


\bibitem[Mialon et~al\mbox{.}(2023)]%
        {mialon2023augmented}
\bibfield{author}{\bibinfo{person}{Gr{\'e}goire Mialon},
  \bibinfo{person}{Roberto Dess{\`\i}}, \bibinfo{person}{Maria Lomeli},
  \bibinfo{person}{Christoforos Nalmpantis}, \bibinfo{person}{Ram Pasunuru},
  \bibinfo{person}{Roberta Raileanu}, \bibinfo{person}{Baptiste Rozi{\`e}re},
  \bibinfo{person}{Timo Schick}, \bibinfo{person}{Jane Dwivedi-Yu},
  \bibinfo{person}{Asli Celikyilmaz}, {et~al\mbox{.}}}
  \bibinfo{year}{2023}\natexlab{}.
\newblock \showarticletitle{Augmented language models: a survey}.
\newblock \bibinfo{journal}{\emph{arXiv preprint arXiv:2302.07842}}
  (\bibinfo{year}{2023}).
\newblock


\bibitem[Newell et~al\mbox{.}(1972)]%
        {newell1972human}
\bibfield{author}{\bibinfo{person}{Allen Newell},
  \bibinfo{person}{Herbert~Alexander Simon}, {et~al\mbox{.}}}
  \bibinfo{year}{1972}\natexlab{}.
\newblock \bibinfo{booktitle}{\emph{Human problem solving}}.
  Vol.~\bibinfo{volume}{104}.
\newblock \bibinfo{publisher}{Prentice-hall Englewood Cliffs, NJ}.
\newblock


\bibitem[Paranjape et~al\mbox{.}(2023)]%
        {paranjape2023art}
\bibfield{author}{\bibinfo{person}{Bhargavi Paranjape}, \bibinfo{person}{Scott
  Lundberg}, \bibinfo{person}{Sameer Singh}, \bibinfo{person}{Hannaneh
  Hajishirzi}, \bibinfo{person}{Luke Zettlemoyer}, {and}
  \bibinfo{person}{Marco~Tulio Ribeiro}.} \bibinfo{year}{2023}\natexlab{}.
\newblock \showarticletitle{ART: Automatic multi-step reasoning and tool-use
  for large language models}.
\newblock  (\bibinfo{year}{2023}).
\newblock
\showeprint[arxiv]{2303.09014}~[cs.CL]


\bibitem[Radford et~al\mbox{.}(2018)]%
        {radford2018improving}
\bibfield{author}{\bibinfo{person}{Alec Radford}, \bibinfo{person}{Karthik
  Narasimhan}, \bibinfo{person}{Tim Salimans}, \bibinfo{person}{Ilya
  Sutskever}, {et~al\mbox{.}}} \bibinfo{year}{2018}\natexlab{}.
\newblock \showarticletitle{Improving language understanding by generative
  pre-training}.
\newblock  (\bibinfo{year}{2018}).
\newblock


\bibitem[Reimers and Gurevych(2019)]%
        {reimers2019sentence}
\bibfield{author}{\bibinfo{person}{Nils Reimers} {and} \bibinfo{person}{Iryna
  Gurevych}.} \bibinfo{year}{2019}\natexlab{}.
\newblock \showarticletitle{Sentence-bert: Sentence embeddings using siamese
  bert-networks}.
\newblock \bibinfo{journal}{\emph{arXiv preprint arXiv:1908.10084}}
  (\bibinfo{year}{2019}).
\newblock


\bibitem[Rosenbloom et~al\mbox{.}(2016)]%
        {rosenbloom2016sigma}
\bibfield{author}{\bibinfo{person}{Paul~S Rosenbloom}, \bibinfo{person}{Abram
  Demski}, {and} \bibinfo{person}{Volkan Ustun}.}
  \bibinfo{year}{2016}\natexlab{}.
\newblock \showarticletitle{The Sigma cognitive architecture and system:
  Towards functionally elegant grand unification}.
\newblock \bibinfo{journal}{\emph{Journal of Artificial General Intelligence}}
  \bibinfo{volume}{7}, \bibinfo{number}{1} (\bibinfo{year}{2016}),
  \bibinfo{pages}{1}.
\newblock


\bibitem[Schick et~al\mbox{.}(2023)]%
        {schick2023toolformer}
\bibfield{author}{\bibinfo{person}{Timo Schick}, \bibinfo{person}{Jane
  Dwivedi-Yu}, \bibinfo{person}{Roberto Dess{\`\i}}, \bibinfo{person}{Roberta
  Raileanu}, \bibinfo{person}{Maria Lomeli}, \bibinfo{person}{Luke
  Zettlemoyer}, \bibinfo{person}{Nicola Cancedda}, {and}
  \bibinfo{person}{Thomas Scialom}.} \bibinfo{year}{2023}\natexlab{}.
\newblock \showarticletitle{Toolformer: Language models can teach themselves to
  use tools}.
\newblock \bibinfo{journal}{\emph{arXiv preprint arXiv:2302.04761}}
  (\bibinfo{year}{2023}).
\newblock


\bibitem[Shum et~al\mbox{.}(2023)]%
        {shum2023automatic}
\bibfield{author}{\bibinfo{person}{KaShun Shum}, \bibinfo{person}{Shizhe Diao},
  {and} \bibinfo{person}{Tong Zhang}.} \bibinfo{year}{2023}\natexlab{}.
\newblock \showarticletitle{Automatic Prompt Augmentation and Selection with
  Chain-of-Thought from Labeled Data}.
\newblock  (\bibinfo{year}{2023}).
\newblock
\showeprint[arxiv]{2302.12822}~[cs.CL]


\bibitem[Touvron et~al\mbox{.}(2023)]%
        {touvron2023llama}
\bibfield{author}{\bibinfo{person}{Hugo Touvron}, \bibinfo{person}{Thibaut
  Lavril}, \bibinfo{person}{Gautier Izacard}, \bibinfo{person}{Xavier
  Martinet}, \bibinfo{person}{Marie-Anne Lachaux},
  \bibinfo{person}{Timoth{\'e}e Lacroix}, \bibinfo{person}{Baptiste
  Rozi{\`e}re}, \bibinfo{person}{Naman Goyal}, \bibinfo{person}{Eric Hambro},
  \bibinfo{person}{Faisal Azhar}, {et~al\mbox{.}}}
  \bibinfo{year}{2023}\natexlab{}.
\newblock \showarticletitle{Llama: Open and efficient foundation language
  models}.
\newblock \bibinfo{journal}{\emph{arXiv preprint arXiv:2302.13971}}
  (\bibinfo{year}{2023}).
\newblock


\bibitem[Wang et~al\mbox{.}(2023)]%
        {wang2023selfconsistency}
\bibfield{author}{\bibinfo{person}{Xuezhi Wang}, \bibinfo{person}{Jason Wei},
  \bibinfo{person}{Dale Schuurmans}, \bibinfo{person}{Quoc Le},
  \bibinfo{person}{Ed Chi}, \bibinfo{person}{Sharan Narang},
  \bibinfo{person}{Aakanksha Chowdhery}, {and} \bibinfo{person}{Denny Zhou}.}
  \bibinfo{year}{2023}\natexlab{}.
\newblock \showarticletitle{Self-Consistency Improves Chain of Thought
  Reasoning in Language Models}.
\newblock  (\bibinfo{year}{2023}).
\newblock
\showeprint[arxiv]{2203.11171}~[cs.CL]


\bibitem[Wang et~al\mbox{.}(2022)]%
        {wang2022self}
\bibfield{author}{\bibinfo{person}{Xuezhi Wang}, \bibinfo{person}{Jason Wei},
  \bibinfo{person}{Dale Schuurmans}, \bibinfo{person}{Quoc Le},
  \bibinfo{person}{Ed Chi}, {and} \bibinfo{person}{Denny Zhou}.}
  \bibinfo{year}{2022}\natexlab{}.
\newblock \showarticletitle{Self-consistency improves chain of thought
  reasoning in language models}.
\newblock \bibinfo{journal}{\emph{arXiv preprint arXiv:2203.11171}}
  (\bibinfo{year}{2022}).
\newblock


\bibitem[Wei et~al\mbox{.}(2023)]%
        {wei2023chainofthought}
\bibfield{author}{\bibinfo{person}{Jason Wei}, \bibinfo{person}{Xuezhi Wang},
  \bibinfo{person}{Dale Schuurmans}, \bibinfo{person}{Maarten Bosma},
  \bibinfo{person}{Brian Ichter}, \bibinfo{person}{Fei Xia},
  \bibinfo{person}{Ed Chi}, \bibinfo{person}{Quoc Le}, {and}
  \bibinfo{person}{Denny Zhou}.} \bibinfo{year}{2023}\natexlab{}.
\newblock \showarticletitle{Chain-of-Thought Prompting Elicits Reasoning in
  Large Language Models}.
\newblock  (\bibinfo{year}{2023}).
\newblock
\showeprint[arxiv]{2201.11903}~[cs.CL]


\bibitem[Yang et~al\mbox{.}(2022)]%
        {yang2022seqzero}
\bibfield{author}{\bibinfo{person}{Jingfeng Yang}, \bibinfo{person}{Haoming
  Jiang}, \bibinfo{person}{Qingyu Yin}, \bibinfo{person}{Danqing Zhang},
  \bibinfo{person}{Bing Yin}, {and} \bibinfo{person}{Diyi Yang}.}
  \bibinfo{year}{2022}\natexlab{}.
\newblock \showarticletitle{SeqZero: Few-shot Compositional Semantic Parsing
  with Sequential Prompts and Zero-shot Models}.
\newblock  (\bibinfo{year}{2022}).
\newblock
\showeprint[arxiv]{2205.07381}~[cs.CL]


\bibitem[Yao et~al\mbox{.}(2022)]%
        {yao2022react}
\bibfield{author}{\bibinfo{person}{Shunyu Yao}, \bibinfo{person}{Jeffrey Zhao},
  \bibinfo{person}{Dian Yu}, \bibinfo{person}{Nan Du}, \bibinfo{person}{Izhak
  Shafran}, \bibinfo{person}{Karthik Narasimhan}, {and} \bibinfo{person}{Yuan
  Cao}.} \bibinfo{year}{2022}\natexlab{}.
\newblock \showarticletitle{React: Synergizing reasoning and acting in language
  models}.
\newblock \bibinfo{journal}{\emph{arXiv preprint arXiv:2210.03629}}
  (\bibinfo{year}{2022}).
\newblock


\bibitem[Zelikman et~al\mbox{.}(2022)]%
        {zelikman2022star}
\bibfield{author}{\bibinfo{person}{Eric Zelikman}, \bibinfo{person}{Yuhuai Wu},
  \bibinfo{person}{Jesse Mu}, {and} \bibinfo{person}{Noah~D. Goodman}.}
  \bibinfo{year}{2022}\natexlab{}.
\newblock \showarticletitle{STaR: Bootstrapping Reasoning With Reasoning}.
\newblock  (\bibinfo{year}{2022}).
\newblock
\showeprint[arxiv]{2203.14465}~[cs.LG]


\bibitem[Zhang et~al\mbox{.}(2022)]%
        {zhang2022automatic}
\bibfield{author}{\bibinfo{person}{Zhuosheng Zhang}, \bibinfo{person}{Aston
  Zhang}, \bibinfo{person}{Mu Li}, {and} \bibinfo{person}{Alex Smola}.}
  \bibinfo{year}{2022}\natexlab{}.
\newblock \showarticletitle{Automatic Chain of Thought Prompting in Large
  Language Models}.
\newblock  (\bibinfo{year}{2022}).
\newblock
\showeprint[arxiv]{2210.03493}~[cs.CL]


\end{thebibliography}

\appendix

\section{Appendix}

\subsection{Prompts}
% 目前模型对字符的识别非常差
In this section, we mainly put the related Prompts design in the experiment, including the prompts of generating question types as shown in Table \ref{pro:qt} and thinking templates as shown in Table \ref{pro:tem}.

% \paragraph{ \bf The Prompts of generating question type.}
% % 

{\renewcommand{\arraystretch}{1.0}
\begin{table*}[ht]
 % \vspace{-0.67cm}
 \caption{the prompts of generating question type.}
 \label{pro:qt}
 \setlength{\tabcolsep}{1.0mm}
 % \centering
 \small
 
 %\footnotesize
  % \begin{tabular}{lp{\textwidth}} 
  
  \begin{tabularx}{\textwidth}{X}
  \toprule
        \bf E-KAR datasets:  \\
        You are the examiner of the Chinese Civil Service Examination,
        and you need to judge the specific question types of the following analogy questions and don't give an explanation. \\
        Question: \{question\} \\
        Answer: The output must only be in a strict JSON format: "task\_type": "question type". \\
        \midrule 
        \bf Math datasets:  \\
        As a mathematics professor, you need to judge the type of the following question and don't give an explanation \\
        Question: \{question\} \\
        Answer: The output must only be in a strict JSON format: "task\_type": "question type". \\

\bottomrule
% \hline
% \end{tabular} 
\end{tabularx}
\end{table*}
}

{\renewcommand{\arraystretch}{1.0}
\begin{table*}[ht]
 % \vspace{-0.67cm}
 \caption{the main prompts of some thinking templates.}
 \label{pro:tem}
 \setlength{\tabcolsep}{1.0mm}
 % \centering
 \small
 
 %\footnotesize
  % \begin{tabular}{lp{\textwidth}} 
  
  \begin{tabularx}{\textwidth}{X}
  \toprule
        \bf Analogical Thinking (AT):  \\
        For the problem of analogical reasoning, it is completed in three steps.\\
        First conduct an inductive analysis of the given sample data, considering the similarity of the relationship between words; Next, judge whether the sample to be selected is satisfied; Finally check the validity of the mapping and explain if the mapping is correct. \\
        \midrule 
        \bf Decomposition Thinking: \\
         
         1) DT: The following questions can be disassembled into multiple sub-questions to solve, the steps and answers of each sub-question are given, and finally the answer to the following question is given. \\
         2) DST: Disassemble the following complex problems to solve them step by step \\
        \midrule 
        \bf Plan Thinking (PT): \\
        Think carefully about the problem to be solved and make a detailed plan to solve it. \\
        \midrule 
        \bf Step Thinking (ST): \\
        Let's think step by step. \\
\bottomrule
% \hline
% \end{tabular} 
\end{tabularx}
\end{table*}
}

\subsection{Cognitive Architecture}
% 本文的框架结构重点参考文章，下面将对文章列出的重点模块进行详细阐述。
The framework structure of this paper mainly refers to the article \cite{kotseruba202040}, and the key modules listed in the article will be described in detail below.

\paragraph{\bf Attention.} The attention section of the essay primarily focuses on the selective mechanisms utilized by cognitive architectures for prioritizing relevant information and filtering out extraneous data. The main idea is to explore how attentional processes help individuals efficiently allocate cognitive resources to specific stimuli or tasks, as well as to discuss the techniques and models applied to modulate attention in various contexts. In LLM, the primary objective of attention is to interpret the input prompt and discern the intent underlying the words.

\paragraph{\bf Action Selection.} In the action selection section, the main idea revolves around examining the decision-making mechanisms employed by cognitive architectures to choose appropriate actions in response to external stimuli or internal states. This part covers key computational models, methods, and algorithmic logic responsible for determining and selecting goal-directed actions based on the available information and environmental context.

\paragraph{\bf Memory.} The memory section of the essay explores the concepts and models related to the storage and retrieval of information within cognitive architectures. The main idea is to investigate the fundamental features of short-term (or working) and long-term memory, as well as to present the mechanisms by which cognitive systems encode, maintain, and retrieve important information and experiences. We focus on reinforcing the existing knowledge that the model has not yet firmly established. Memory stores the consolidated information in external libraries, effectively functioning as long-term memory for the model.

\paragraph{\bf Learning.} The learning section delves into the processes that help cognitive architectures acquire new information, adapt to new situations, and generalize from previous experiences. The main idea is to examine the different learning paradigms, such as supervised, unsupervised, and reinforcement learning, and their applications in equipping cognitive architectures with the ability to modify and improve their knowledge structures, representations, and decision-making processes. By updating the few-shot content in the prompt, this task can be easily accomplished. This paper achieves learning in the Large Language Model (LLM) by updating the note library, allowing the model to acquire new knowledge and adapt to new information.

\paragraph{\bf Reasoning.} The reasoning section of the essay emphasizes the cognitive processes underlying problem-solving, decision-making, and inference within cognitive architectures. The main idea is to present various approaches and models that demonstrate logical and probabilistic reasoning, as well as to discuss the mechanisms for generating predictions, explanations, and strategies based on available information and knowledge. The module incorporates various templates that enable the model to approach problem-solving situations more effectively and generate well-structured solutions.

\subsection{Examples}
\begin{figure*}[ht]
  \centering
  \includegraphics[width=1\linewidth]{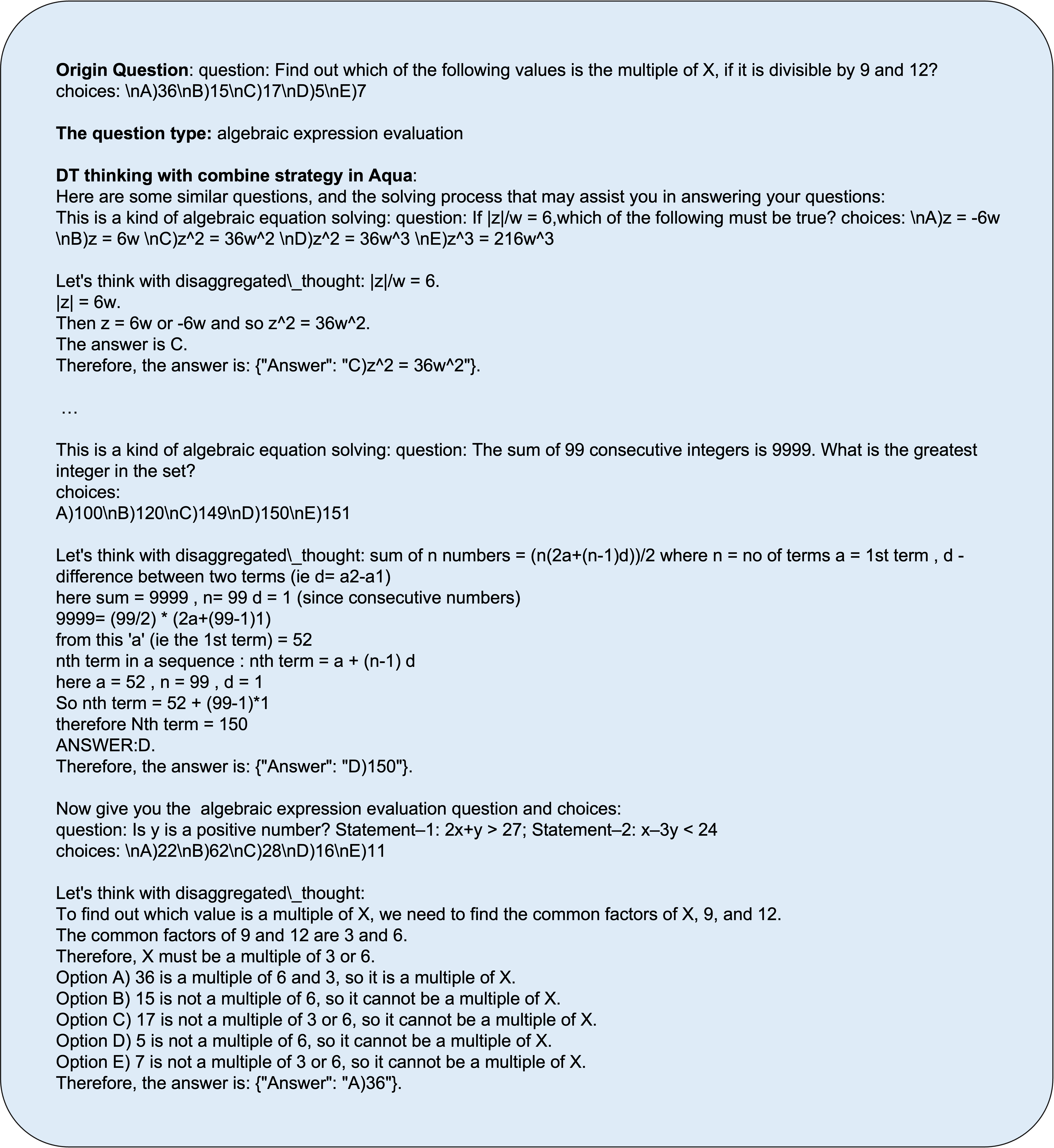}
  \caption{ one example contents for \textbf{AQuA}.}
  % \Description{A figure shows the overall structure of the OlaGPT model.}
  \label{fig:notes_aqua_3}
\end{figure*}

\begin{figure*}[ht]
  \centering
  \includegraphics[width=1\linewidth]{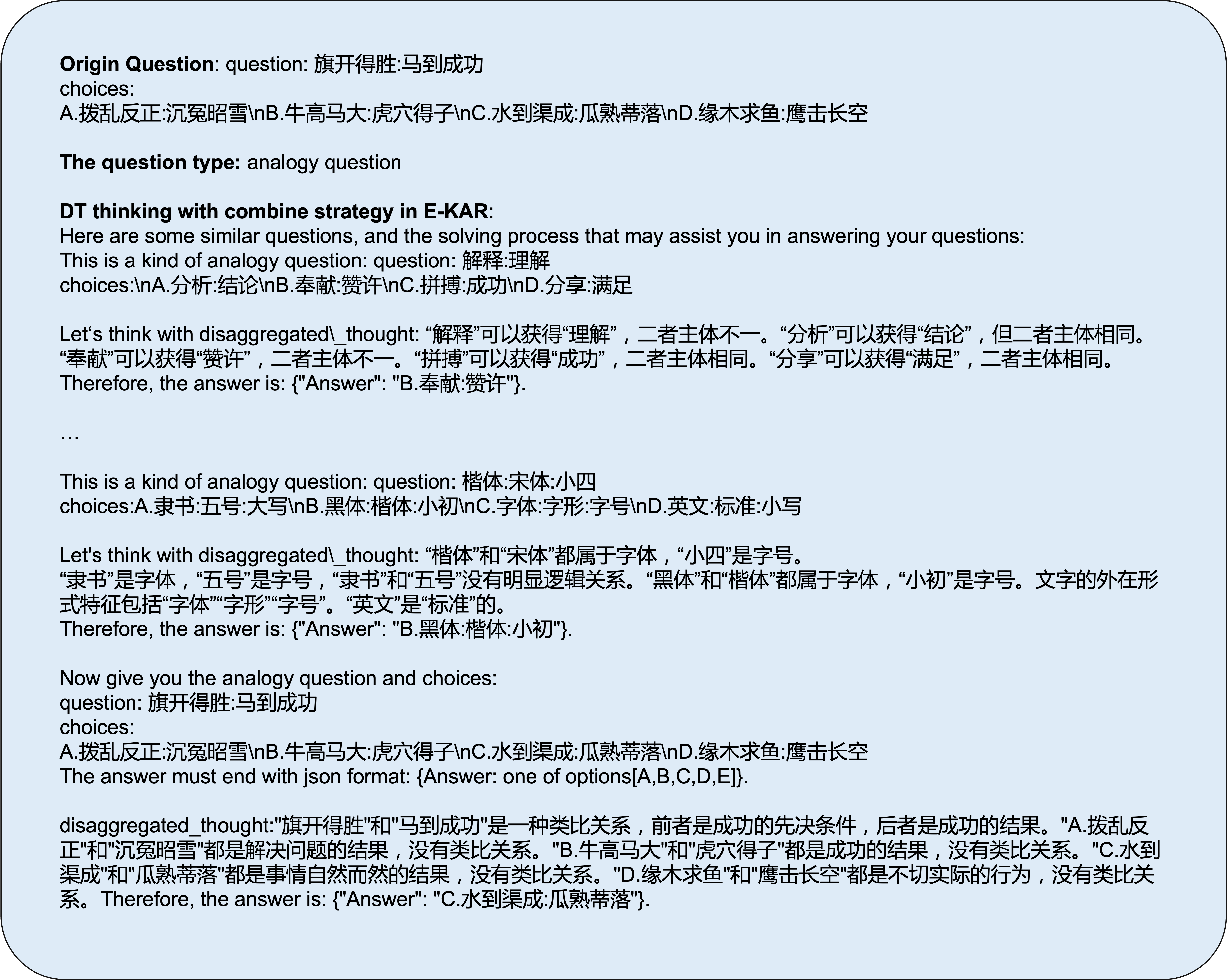}
  \caption{ one example contents for \textbf{E-KAR}.}
  % \Description{A figure shows the overall structure of the OlaGPT model.}
  \label{fig:notes_ekar_3}
\end{figure*}

 % 本小节主要介绍给出具体的执行文本样例，In the Figure \ref{fig:notes_aqua_3, fig:notes_ekar_3}，我们给出了不同数据集在对大模型进行提问时利用combine策略检索到的notes之后最终得到答案的全文本内容。
This section mainly introduces the specific execution text samples. In the Fig \ref{fig:notes_aqua_3} and \ref{fig:notes_ekar_3}, we give the full-text content in different datasets using the $combine$ strategy to retrieve the notes when asking questions about LLMs(gpt-turbo-3.5) with the DT thinking template.

% \subsection{Experimental Details}

\subsection{Templates analysis}
% 模板之间的分析结果
The performance of each template is presented in Table \ref{tab:all_result}. According to Table \ref{tab:template_precision}, there are fluctuations in each template when executed on different datasets, which suggests that utilizing model ensembling may lead to improved performance. Consequently, this work employs a voting mechanism to capitalize on the strengths of various templates while addressing their individual limitations.

On the \textbf{AQuA} dataset, the best-performing template is DT, which demonstrates the highest average accuracy, reaching 0.5807. In the \textbf{E-KAR} Chinese dataset, the top-performing template is ST, achieving an accuracy of 0.4015. These results highlight the effectiveness of using different templates tailored to the specificities of each dataset in order to maximize performance.
{
\renewcommand{\arraystretch}{1.2}
\begin{table*}
	\caption{Precision analysis of the thinking templates.}
	\label{tab:template_precision}
	\centering
	\begin{tabular}{@{}c|c|c|c|c|c|c|c|c@{}}
		\toprule
		%\hline
		\multirow{2}{*}{} &\multicolumn{4}{c|}{AQuA}&\multicolumn{4}{c}{E-KAR (Chinese)}   \\
		\cline{2-9} & zero-shot & random & retrieval & combine & zero-shot & random &  retrieval & combine  \\
		\midrule 
            Range & 0.0512 & 0.0630 & 0.0275 & 0.0433 & 0.0478 & 0.0627 & 0.0508 & 0.0507 \\
            Mean & 0.5345 & 0.5590 & 0.5935 & 0.5768 & 0.3648 & 0.3940 & 0.4036 & 0.4042 \\
		\midrule
		\bottomrule
	\end{tabular}
\end{table*}
}

\subsection{Vote analysis}
% 探索准确率峰值
In this study, we investigate voting methods in model ensembles, employing two distinct approaches for fusing the results derived from different models. The first approach entails extracting candidate answers using regular expressions and selecting the most frequently occurring answer as the ensemble's output. The second approach feeds the predicted outputs and their analyses from diverse models into GPT-3.5-turbo, which subsequently generates the final answer.

Considering the voting method utilizing regular expressions for answer extraction, a supremum and infimum inherently exist. To clarify this, the study introduces two metrics evaluating the answers produced by various models for identical questions: accuracy and incorrect consistency. Accuracy represents the proportion of models delivering the correct answer, while incorrect consistency refers to the proportion associated with the most frequently recurring answer, excluding the correct one.

The supremum corresponds to the proportion of questions where accuracy either surpasses or equals consistency. In contrast, the infimum signifies the proportion where accuracy exceeds consistency. The GPT-3.5-turbo-based voting method results reveal an average increase of 0.0561 and 0.0366 in comparison to the infimum derived through regular expression answer extraction for the \textbf{AQuA} and \textbf{E-KAR} (Chinese) datasets, respectively. When compared to the supremum, accuracy remains an average of 0.0325 lower for the \textbf{AQuA} dataset and 0.0485 lower for the \textbf{E-KAR} (Chinese) dataset. Consequently, the outcomes acquired through LLM voting exhibit a higher degree of robustness.

\begin{figure*}[htbp]
  \centering
  \includegraphics[width=1\linewidth]{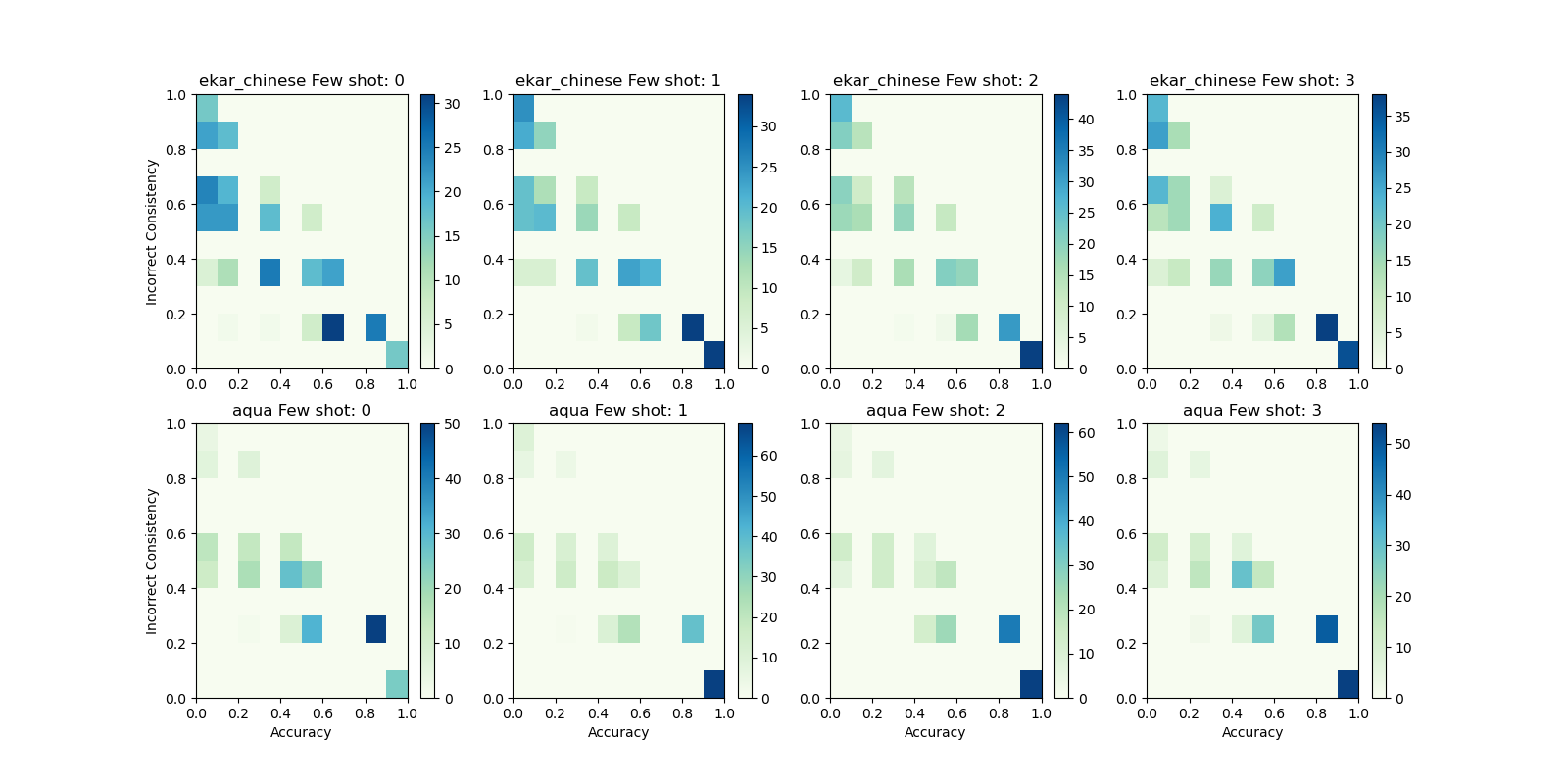}
  \caption{ The Accuracy and Incorrect Consistency of \textbf{AQuA} and \textbf{E-KAR} (Chinese).}
  % \Description{A figure shows the overall structure of the OlaGPT model.}
  \label{fig:heatmap_res}
\end{figure*}

Furthermore, the heatmap shown in Fig \ref{fig:heatmap_res} demonstrates that answers derived from different templates tend to cluster, indicating that these templates produce analogous judgments for the same questions.

{\renewcommand{\arraystretch}{1.2}
\begin{table*}[htbp]
   
	\caption{Explore the high accuracy of the theory.}
	\label{tab:high}
	\centering
% 	\footnotesize
	\begin{tabular}{@{}c|c|c|c|c|c|c|c|c@{}}
		\toprule
		%\hline
		\multirow{2}{*}{Combination} &\multicolumn{4}{c|}{AQuA}&\multicolumn{4}{c}{E-KAR (Chinese)}  \\
		\cline{2-9} & zero-shot & $random$ & $retrieval$ & $combine$ & zero-shot & $random$ & $retrieval$ & $combine$ \\
		\midrule 
            regex-upper & 0.6457 & 0.6614 & 0.7047 & 0.7283 & 0.4537 & 0.5015 & 0.4866 & 0.4806 \\
            regex-lower & 0.5315 & 0.5906 & 0.6614 & 0.6024 & 0.3552 & 0.4179 & 0.4030 & 0.4060 \\
            llm-vote & 0.5984 & 0.6417 & 0.6654 & 0.7047 & 0.4000 & 0.4418 & 0.4358 & 0.4507 \\
            reg-vote &  0.5945 & 0.6496 & 0.6732 & 0.6772 & 0.4209 & 0.4597 & 0.4567 & 0.4716 \\
            
		\midrule
		\bottomrule
	\end{tabular}
\end{table*}
}

\end{document}